\documentclass[10pt,conference]{IEEEtran}
\ifCLASSOPTIONcompsoc
  \usepackage[nocompress]{cite}
\else
  \usepackage{cite}
\fi
\ifCLASSINFOpdf
   \usepackage[pdftex]{graphicx}
  \graphicspath{{./figures/}}
   \DeclareGraphicsExtensions{.pdf,.jpeg,.png,.svg}
\else
   \usepackage[dvips]{graphicx}
\fi
\usepackage{amsmath}
\usepackage{amssymb}
\DeclareMathOperator*{\argmax}{arg\,max}
\DeclareMathOperator*{\argmin}{arg\,min}
\usepackage{float}
\usepackage{algorithmic}
\usepackage{algorithm}

\usepackage{xcolor}
\ifCLASSOPTIONcompsoc
  \usepackage[caption=false,font=footnotesize,labelfont=sf,textfont=sf]{subfig}
\else
  \usepackage[caption=false,font=footnotesize]{subfig}
\fi
\usepackage{amsthm}
\newtheorem{theorem}{Theorem}[section]

\newtheorem{lemma}[theorem]{Lemma}

\usepackage{multirow}
\begin{document}
\title{Avoiding Communication in Logistic Regression}

\author{
\IEEEauthorblockN{Aditya Devarakonda}
\IEEEauthorblockA{Institute for Data Intensive Engineering and Science\\
Johns Hopkins University\\Baltimore, Maryland\\adi@jhu.edu}
\and
\IEEEauthorblockN{James Demmel}
\IEEEauthorblockA{Department of EECS\\Department of Mathematics\\
University of California, Berkeley\\Berkeley, California\\demmel@berkeley.edu}
}

%<-this % stops a space


\markboth{Avoid. Comm. Logistic Regression}%
{}



\IEEEtitleabstractindextext{%
\begin{abstract}
Stochastic gradient descent (SGD) is one of the most widely used optimization methods for solving various machine learning problems. SGD solves an optimization problem by iteratively sampling a few data points from the input data, computing gradients for the selected data points, and updating the solution. However, in a parallel setting, SGD requires interprocess communication at every iteration. We introduce a new communication-avoiding technique for solving the logistic regression problem using SGD. This technique re-organizes the SGD computations into a form that communicates every $s$ iterations instead of every iteration, where $s$ is a tuning parameter. We prove theoretical flops, bandwidth, and latency upper bounds for SGD and its new communication-avoiding variant. Furthermore, we show experimental results that illustrate that the new Communication-Avoiding SGD (CA-SGD) method can achieve speedups of up to $4.97\times$ on a high-performance Infiniband cluster without altering the convergence behavior or accuracy.
\end{abstract}
\begin{IEEEkeywords}
Communication Avoidance, Logistic Regression, Stochastic Gradient Descent, Binary Classification.
\end{IEEEkeywords}}

\maketitle

\IEEEdisplaynontitleabstractindextext

% \fi
\IEEEpeerreviewmaketitle

%\IEEEraisesectionheading{\section{Introduction}\label{sec:introduction}}
\section{Introduction}\label{sec:introduction}

%!TEX ROOT=../calogistic-tpds.tex
%\textcolor{red}{\textbf{To be completed.}}
\IEEEPARstart{O}{ptimization} methods are at the core of many machine learning applications. For example, the areas of computer vision and natural language processing make use of large machine learning models that have been trained on vast amounts data to enable computers to automatically classify images or translate speech to text. Much of the prediction power is derived from solving nonlinear optimization problems which often perform regression (linear and nonlinear) or classification (binary and multiclass). In order to solve these optimization problems, we often compute first- or second-order derivatives and incrementally update the solution until it converges. Such approaches to solving optimization problems \cite{wright15, bottou10, nesterov12} have been well-studied, however, with the rise of multi-core/multi-node processing these optimization methods must now be parallelized across multiple cores/nodes. As a result, studying and improving the parallel performance of these optimization methods is imperative and would impact many application areas.

 In this paper, we will focus on the stochastic gradient descent (SGD) method \cite{bottou10} for solving binary classification problems using the logistic regression model. SGD solves the logistic regression problem by iteratively sampling a few data points from the input data, computing the gradient of the logistic loss function, and updating the solution. As a result, parallel variants of SGD require interprocess communication at every iteration. On modern computing hardware, where communication cost often dominates computation cost, the running time of parallel SGD is often dominated by communication cost.

We model communication cost on a distributed-memory parallel cluster in terms of two costs: latency and bandwidth. Our goal is to show that the latency cost of SGD, which is often the dominant cost, can be improved by a tunable factor of $s$ by trading a factor of $sb$ additional bandwidth and computation, where $b$ is a tunable batch size. %These results are obtained by applying a communication-avoiding technique, which is inspired by related work in Communication-Avoiding Krylov Subspace methods \cite{hoemmen, carson, chronopoulos89, chronopoulosthesis, mohiyuddin09, demmel08}.
\textit{The main contributions of this paper are}:
\begin{itemize}
\item Derivation of a Communication-Avoiding SGD (CA-SGD) method for solving the logistic regression problem which reduces SGD latency cost by a tunable factor of $s$ in exchange for a factor of $sb$ additional bandwidth and computation.
\item Theoretical analysis of the flops, bandwidth, and latency costs of SGD and CA-SGD under two input matrix partitioning schemes: 1D-block row and 1D-block column.
\item Numerical experiments which illustrate that CA-SGD is numerically stable for very large values of $s$.
\item Performance experiments which illustrate that CA-SGD can attain speedups of up to $4.97\times$ over SGD and can scale out to $4\times$ as many cores.
\end{itemize}
\subsection{Logistic Regression} 
Logistic regression is a supervised learning model used to predict the probability of data points belonging to one of two classes (binary classification). This model is widely used in many applications like predicting disease risk, website click-through prediction, and fraud detection which often require classification of data in terms of two classes. 

We will now briefly derive the optimization problem for logistic regression used in this paper. We begin by defining the logistic function:
\begin{align*}
\sigma(\theta) = \frac{e^{\theta}}{1 + e^{\theta}} \equiv \frac{1}{1 + e^{-\theta}}. 
\end{align*}
Now suppose we are given a dataset $A \in \mathbb{R}^{m \times n}$ with $m$ data points (rows of $A$) and $n$ features (columns of $A$) and a vector of labels (one label per data point), $y \in \mathbb{R}^{m}$ such that $y_i \in \{-1, +1\}~\forall~i = 1,\ldots,m$. Given such a dataset, the goal is to compute a vector $x \in \mathbb{R}^n$ of weights for each feature that maximizes the probability of correctly classifying the input data. Using the logistic function, we can model the probability for each data point as
\begin{align}
P(y_i | a_ix) = 
\begin{cases}\sigma(a_ix) & y_i = +1 \\
1 - \sigma(a_ix) & y_i = -1\label{lrprob},
\end{cases}
\end{align}
where $a_i$ is the $i$-th data point (row) in $A$ and $x$ is the unknown vector of weights. In this paper we assume the labels are $-1$ and $+1$ because this label choice leads to fewer terms in the optimization problem. This results in a more concise derivation of communication-avoiding SGD (CA-SGD). Our results also hold for other mathematically equivalent formulations (i.e. labels that are $0$ and $+1$, etc.) of the logistic regression problem.

Since $1 - \sigma(a_ix) = \sigma(-a_ix)$ by symmetry of the logistic function, \eqref{lrprob} can be further simplified to, 
\begin{align*}
P(y_i | a_i x) = \sigma(y_ia_i x).
\end{align*}
From this the optimization problem can be defined as
\begin{align}
\argmax_x \prod_{i = 1}^m \sigma(y_ia_ix) = \argmax_x \prod_{i=1}^m \frac{1}{1 + \exp(-y_ia_ix)}\label{lrlike},
\end{align}
which computes the weghts, $x$, that maximizes the likelihood function. Finally, by taking the negative log of the likelihood, we can cast \eqref{lrlike} into the form of empirical risk minimization,
\begin{align}
\argmin_x  &~F(A, x, y),\label{eq:lr}\\
\text{where}~F(A, x, y) &= \frac{1}{m}\sum_{i = 1}^m \log{(1 + \exp\left(-y_ia_ix\right))}\nonumber.
\end{align}
Unlike linear regression, \eqref{eq:lr} does not have a closed-form solution and cannot be solved using direct methods (i.e. through matrix factorization). However, one approach to solving this problem is to update the solution iteratively using the gradient of \eqref{eq:lr} until the solution converges. The gradient with respect to $x$ is given by
\begin{align}
\nabla F(A,x,y) = \frac{1}{m}\sum_{i = 1}^m \frac{-\tilde a_i^T}{1 + \exp\left(\tilde a_ix\right)},\label{eq:lrgrad}
\end{align}
where $\tilde a_i = y_ia_i~\forall~i=1,\ldots,m$ (i.e. the rows of $A$ scaled by their corresponding labels). In matrix form we will use the notation $\tilde A = A \circ y$, where $\circ$ represents scaling the $i$-th row of $A$ by the $i$-th element of $y$. For convenience we can rewrite \eqref{eq:lrgrad} in matrix form as
\begin{align}
\nabla F(A,x,y)  = -\frac{1}{m}\tilde A^T \left(\vec{1} \oslash \left(\vec{1} + \exp\left(\tilde Ax\right)\right) \right)~,\label{eq:matrixlrgrad}
\end{align}
where $\oslash$ is the elementwise division operation and $\exp(\cdot)$ is now the exponential function applied elementwise to the vector $\tilde Ax$. For clarity we will use the notation $sig\left(\tilde A x\right) = \vec{1} \oslash \left(\vec{1} + \exp\left(\tilde Ax\right)\right)$, which is the sigmoid function applied to the vector $\tilde A x$. We can then rewrite \eqref{eq:matrixlrgrad} as 
\begin{align}
\nabla F(A,x,y)  = -\frac{1}{m}\tilde A^T sig\left(\tilde A x\right)~.\label{eq:matrixlrgradsimp}
\end{align}
From \eqref{eq:lrgrad} the solution vector, $x_h$, for iteration $h$ can be obtained by,
\begin{align}
x_h = x_{h-1} - \eta_h \nabla F(A,x_{h-1},y),\label{eq:lrupdate}
\end{align}
where $x_{h-1}$ is the solution vector from the previous iteration and $\eta_h$ is the learning rate (or step size) at iteration $h$ which determines by how much the solution moves in the $\nabla F(A,x_{h-1}, y)$ direction. This is the well-known gradient descent (GD) method for iteratively refining the solution, $x_{h-1}$, until it converges to the optimal solution. $\eta_h$ is a tuning parameter which may drastically affect the convergence behavior of gradient descent. 

Another method often used to solve the logistic regression problem is Stochastic Gradient Descent (SGD). Instead of using the entire matrix $A$ to compute the gradient, $\nabla F(A,x_{h-1}, y)$, SGD computes the gradient for only a subset of the data points (rows) in $A$. Strictly speaking, SGD samples only $1$ row of $A$ at each iteration. However, we will generalize this to a tunable batch size of $b$ rows sampled from the matrix $A$. The resulting update for SGD with batch size $b$ becomes
\begin{align}
x_h = x_{h-1} - \eta_h \nabla F(\mathbb{I}_hA, x_{h-1}, \mathbb{I}_hy), \label{eq:lrsgdupdate}
\end{align}
where $\mathbb{I}_h$ is a matrix in $\mathbb{R}^{b \times m}$ which corresponds to $b$ rows sampled uniformly at random without replacement from the identity matrix, $I_m$. As a result, $\mathbb{I}_hA$ and $\mathbb{I}_hy$ simply select $b$ rows of $A$ and their corresponding $b$ labels from $y$. Note that if $b = m$, SGD is equivalent to gradient descent except with the rows of $A$ permuted every iteration. The resulting SGD algorithm is shown in Algorithm \ref{alg:sgd}.
\begin{algorithm}[H]
\caption{Stochastic Gradient Descent}
\label{alg:sgd}
\begin{algorithmic}[1]
%\floatname{algorithm}{Procedure}
%\Procedure{Gradient Descent}
\REQUIRE{$A \in \mathbb{R}^{m \times n}, y \in \mathbb{R}^{m}, \eta_0, b, H'.$}
\ENSURE{$x_H \in \mathbb{R}^n$}
\STATE{$x_0 = \vec{0},~\tilde A = A \circ y$}
\FOR{$h = 1, 2\ldots H'$}
\STATE {choose $\{i_k \in [m]| k = 1, 2, \ldots, b\}$ uniformly at random without replacement.}
\STATE {$\mathbb{I}_h = [e_{i_1}, e_{i_2}, \ldots, e_{i_k}, \ldots, e_{i_b}]^T$} where $e_{i_k} \in \mathbb{R}^m$ is the $k$-th standard basis vector.
\STATE {$x_h = x_{h-1} + \frac{\eta_h}{m}\tilde A^T\mathbb{I}_h^T sig\left(\mathbb{I}_h\tilde A x_{h-1}\right)$}
\ENDFOR
\RETURN $x_{H'}$
%\caption{Gradient Descent}
\end{algorithmic}
\end{algorithm}
%Figure \ref{fig:sgdobjeta} shows a comparison of the convergence behavior of SGD with $b = 1$ for various values of $\eta_h$ for $1,000$ epochs on the a6a dataset. Note that one epoch of SGD requires $m$ iterations\footnote{We do not require that SGD touch all rows of $A$ in one epoch. Only that $m$ rows of $A$ are sampled.}. Immediately we can see that the values of $\eta_h$ are much higher for SGD than for GD. This is to be expected since we are computing the gradient with respect to just one row of $A$. As with GD, we can obseve that $\eta_h$ must be chosen carefully for SGD. If $\eta_h$ is too high ($\eta_h = 1000$) we observe that convergence becomes slow and oscillates sharply. Oscillation is to be expected with SGD since we are computing gradients using only one row of $A$ at each iteration, which is often not the optimal gradient direction in which to move (when compared to gradient descent). The sharpness of the oscillation, however, is indicative of a learning rate that is too high where we move too far along each gradient direction. If $\eta_h$ is too small, then convergence of SGD is smoother but takes longer to converge. We observe that $\eta_h = 10$ performs the best and exhibits smooth convergence with less oscillation than higher values of $\eta_h$. 

\begin{figure}
\centering
\includegraphics[width=.7\columnwidth, bb=0 0 370 265]{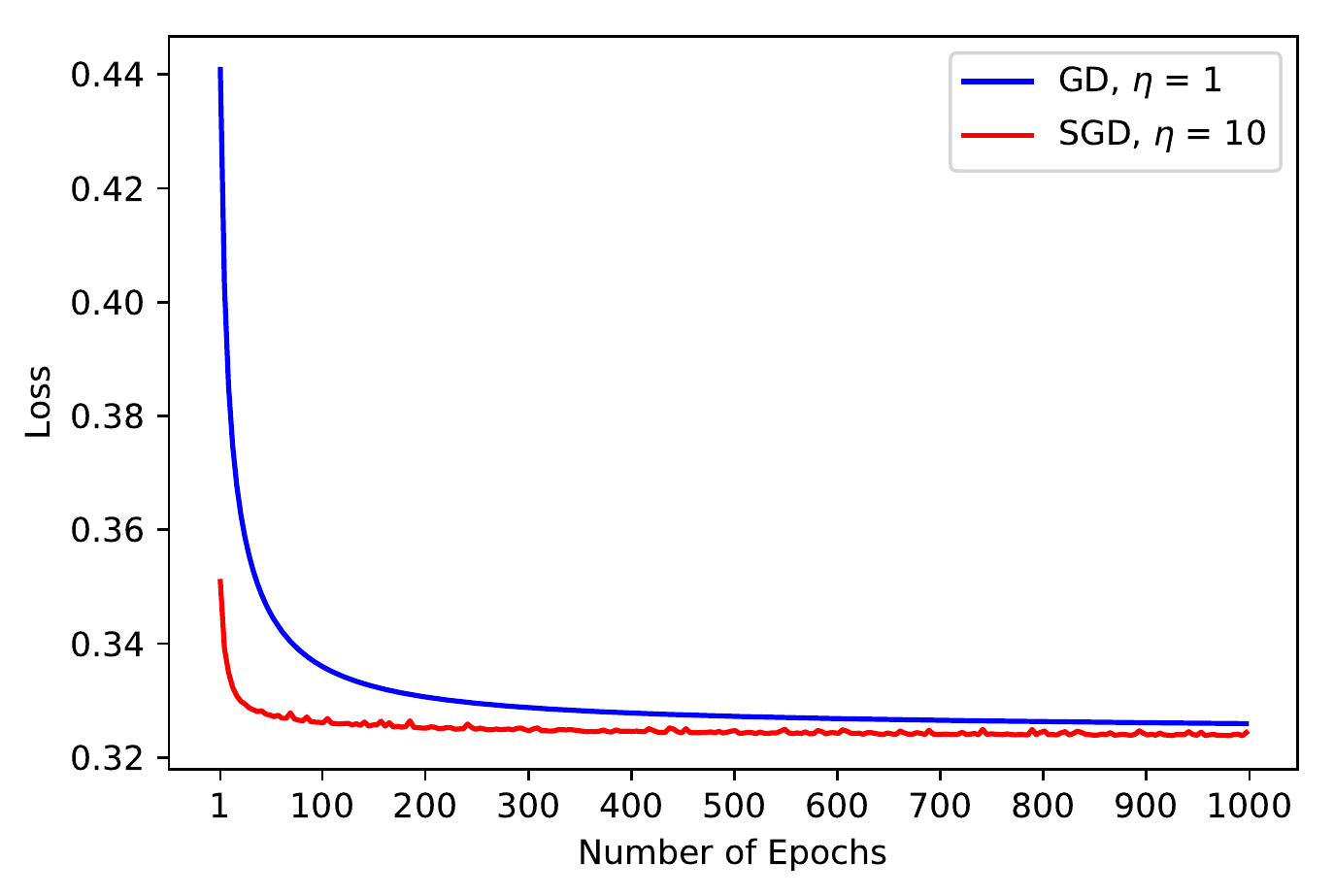}
\caption{Comparison of Gradient Descent (GD, blue) and Stochastic Gradient Descent (SGD, red) loss over $1000$ epochs. We select the best learning rate for GD and SGD through offline tuning.}\label{fig:compareobj}
%\includegraphics[width=\columnwidth]{./figures/a6a_GD_SGD_accuracy_compare_zoom}
%\caption{Comparison of Gradient Descent (GD, blue) and Stochastic Gradient Descent (SGD, red) accuracy over $1000$ epochs corresponding to loss shown in Figure \ref{fig:compareobj}. }\label{fig:compareacc}
\end{figure}

Figure \ref{fig:compareobj} compares the convergence behavior of GD and SGD for the best $\eta_h$ setting on the a6a dataset from the LIBSVM \cite{libsvm} repository. We perform tuning of $\eta_h$ offline and show the best setting for GD and SGD, respectively. Note that many strategies exist for finding optimal, static learning rates and recent results have also illustrated that adaptive learning rates work well for convex optimization methods. In this paper, we focus on introducing the communication-avoiding (CA) derivation and studying its numerical and performance characteristics. We leave the effects of various learning rate strategies on the CA technique for future work.

In Figure \ref{fig:compareobj}, we observe that GD ($\eta_h = 1$) and SGD ($\eta_h = 10$) on the a6a dataset. We can observe that SGD converges faster than GD over the $1000$ epochs. This suggests that SGD is a better choice of algorithm for applications of logistic regression. Furthermore, the fast initial convergence of SGD suggests that it is a much better algorithm if a low-accuracy solution is sufficient.

When $b << m$, each iteration of SGD requires less computation than GD (by a factor of $\frac{m}{b}$) while one epoch of SGD performs the same amount of computation as one iteration (one epoch) of GD. In the distributed-memory parallel setting with $A$ distributed across several processors each iteration of GD and SGD requires communication. Since SGD requires $\frac{m}{b}$ iterations to match GD, SGD requires a factor of $\frac{m}{b}$ more rounds of communication. On modern parallel hardware where communication is often the dominant cost, SGD requires orders of magnitude more communication than GD. This paper focuses on reducing the communication bottleneck in SGD without altering the convergence rate and behavior up to floating-point error.

%\begin{figure}
%\includegraphics[width=.9\columnwidth]{./figures/a6a_SGD_loss_eta}
%\caption{Comparison of Stochastic Gradient Descent loss vs. number of epochs on the LIBSVM a6a dataset ($m = 11,220, n = 123$) for various learning rates. }\label{fig:sgdobjeta}
%\includegraphics[width=\columnwidth]{./figures/a6a_SGD_accuracy_eta}
%\caption{Comparison of Stochastic Gradient Descent loss vs. number of epochs on the LIBSVM a6a dataset ($m = 11,220, n = 123$) for various learning rates.}\label{fig:sgdacceta}
%\end{figure}

\section{Related Work}
Many techniques exist in literature which attempt to reduce the communication bottleneck in machine learning. For example, HOGWILD! \cite{hogwild} uses an asynchronous SGD method for the shared-memory setting where each thread computes gradients and updates the solution vector without synchronization. Due to the lack of synchronization, a thread may overwrite (and undo) the progress another thread has made. Convergence of HOGWILD! is not guaranteed, but will converge with high probability if the solution updates are sufficiently sparse and if there is bounded delay. In HOGWILD!, the latency bottleneck is reduced at the expense of convergence rate.

CoCoA \cite{cocoa, ma17} is a general framework for reducing the synchronization cost of solving various machine learning problems in the distributed-memory setting. CoCoA reduces the synchronization cost by performing coordinate ascent on only the locally stored rows of $A$. After a tunable number of local iterations, the solution from each processor are sum-reduced (or averaged). If too many local iterations are performed, then global convergence will be slow. As with HOGWILD!, the reduction in latency (by defering communication) comes at the expense of convergence rate.
In contrast, our approach does not alter the convergence rate of SGD. Instead, we introduce a tunable communication-avoiding parameter, $s$, that trades off additional computation and bandwidth in order to reduce latency by a factor of $s$. This means that if latency is the dominant cost in SGD, then we can reduce it by a factor of $s$ and attain $s$-fold speedup with our new CA-SGD method.

Our technique is closely related to the one introduced in $s$-step and communcation-avoiding Krylov (CA-Krylov) methods \cite{hoemmen, carson, chronopoulos89, chronopoulosthesis, mohiyuddin09, demmel08}. The $s$-step and CA-Krylov methods work showed that the recurrence relations in Krylov methods can be unrolled by a tunable factor of $s$ and the remaining computation rearranged to avoid synchronization cost in the distributed-memory parallel setting. While the new methods have been shown to be faster, they suffered from numerical instability which subsequent work addressed by introducing techniques to improve numerical stability \cite{carson, Carson13, Carson14, Carson15}.

The same recurrence unrolling technique has been shown to be effective for primal/dual coordinate and block coordinate descent methods and quasi-Newton's method for solving ridge regression, LASSO, and SVM \cite{ppacksvm, soori18, mythesis, siscpaper, ipdpspaper}. This paper extends prior results by illustrating that the technique works for solving the logistic regression problem using SGD where the loss function is nonlinear instead of linear (linear/ridge regression) or piecewise linear (LASSO, SVM).

Unlike CA-Krylov methods, our CA-SGD method does not exhibit any numerical instability even for very large values of $s$. This allows CA-SGD to simply select the value of $s$ that balances the additional computation and bandwidth with the reduction in latency. The prior work on primal and dual block coordinate descent focused on piecewise linear problems. The piecewise linearity ensures that the distributive property can be applied in order to simplify the communication-avoiding derivation. However, this is not true for the logistic regression problem which requires computation of $sig\left(\tilde Ax\right)$ for the gradient. We will show in Section \ref{sec:deriv} that this issue can be managed by making use of additional memory and communication properties of matrix-vector multiply (Lemmas \ref{thm:matvec} and \ref{thm:matvec2}) for the 1D-block column partitioned algorithm. The 1D-block row partitioned algorithm, however, requires an entirely new approach in order to reduce latency by a factor of $s$. This new approach for the 1D-block row partitioned case is described in Section \ref{sec:proofs}.
%The contributions of this paper are:
%\begin{itemize}
%\item Derivation of a communication-avoiding SGD (CA-SGD) method for solving the logistic regression problem.
%\item Algorithms analysis under the distributed-memory parallel setting for two different data paritition schemes that proves that CA-SGD reduces the latency cost by a tunable factor of $s$ at the expense of additional computation and bandwidth.
%\item Numerical stability experiments that illustrate that CA-SGD exhibits the same convergence behavior, convergence rate, and sequence of solutions as the existing SGD method for large values of $s$. 
%\item Performance experiments \textcolor{red}{To be completed.}
%\end{itemize}

%!TEX ROOT=../calogistic-tpds.tex
\begin{algorithm}[t]
\caption{Communication-Avoiding SGD}
\label{alg:casgd}
\begin{algorithmic}[1]
%\floatname{algorithm}{Procedure}
%\Procedure{Gradient Descent}
\REQUIRE{$A \in \mathbb{R}^{m \times n}, y \in \mathbb{R}^{m}, \eta_0, b, s, H'.$}
\ENSURE{$x_H \in \mathbb{R}^n$}
\STATE{$x_0 = \vec{0},~\tilde A = A \circ y$}
\FOR{$h = 0, 2\ldots \frac{H'}{s}$}
	\FOR{$j = 1,2, \ldots s$}
	\STATE {choose $\{i_k \in [m]| k = 1, 2, \ldots, b\}$ uniformly at random without replacement.}
	\STATE {$\mathbb{I}_{sh + j} = [e_{i_1}, e_{i_2}, \ldots, e_{i_k}, \ldots, e_{i_b}]^T$} where $e_{i_k}$ is the $k$-th standard basis vector.
	\ENDFOR
\STATE {Let $Y = \begin{bmatrix}\mathbb{I}_{sh+1}\\ \mathbb{I}_{sh+2}\\ \vdots\\ \mathbb{I}_{sh+s}\end{bmatrix}\tilde A$}
\STATE {$G = YY^T$}
\STATE {$r = Yx_{sh}$}
	\FOR{$j = 1, 2, \ldots s$}
	\STATE {Update $x_{sh+j}$ according to eq. \eqref{eq:caupdate}.}
	%\STATE {$x_{sh+j} = x_{sh-1} + \frac{\eta_{sh+j}}{m} \tilde A^T \mathbb{I}_{sh+j}^T u_{sh+j}$}
	\ENDFOR
\ENDFOR
\RETURN $x_{H'}$
%\caption{Gradient Descent}
\end{algorithmic}
\end{algorithm}
\section{Derivation}\label{sec:deriv}
The Stochastic Gradient Descent (SGD) method is defined by the solution update,
\begin{align*}
x_{h} = x_{h-1} - \eta_{h}\nabla F(\mathbb{I}_h\tilde A,x_h,y),
\end{align*}
where $b$ is the batch size and $\mathbb{I}_h \in \mathbb{R}^{b \times m}$ is a matrix that contains $b$ rows sampled uniformly at random without replacement from the $m$-dimensional identity matrix, $I_m$. We will use the following form in our derivation for the communication avoiding variant,
\begin{align}
x_{h} = x_{h-1} + \frac{\eta_{h}}{m}\tilde A^T \mathbb{I}^T_h sig\left(\tilde Ax_{h-1}\right).\label{eq:xh}
\end{align}
By unrolling the recurrence, we can write $x_{h + 1}$ in terms of $x_{h-1}$,
 \begin{align}
 x_{h+1} &=  x_{h-1} + \frac{\eta_{h}}{m}\tilde A^T \mathbb{I}^T_h sig\left( \mathbb{I}_h\tilde Ax_{h-1}\right) \nonumber\\  &+ \frac{\eta_{h+1}}{m} \tilde A^T \mathbb{I}^T_{h+1}\nonumber \\& sig\left(\mathbb{I}_{h+1}\tilde A x_{h-1} + \frac{\eta_{h}}{m}\mathbb{I}_{h+1}\tilde A\tilde A^T \mathbb{I}^T_h sig\left(\mathbb{I}_h\tilde Ax_{h-1}\right)\right).\label{eq:xh1}
 \end{align}
Note that the term $sig\left(\mathbb{I}_h\tilde A x_{h-1}\right)$ is used twice, once in \eqref{eq:xh} and then again to correct the gradient for $x_{h+1}$ in \eqref{eq:xh1}. Since $sig\left(\mathbb{I}_h\tilde A x_{h-1}\right)$ has already been computed from \eqref{eq:xh}, it can be reused in future solution updates. As a result, the recurrence unrolled solution updates still require only one $sig(\cdot)$ computation per solution update. This is important since the exponential operation is more expensive than typical arithmetic operations.

For convenience we will change the loop iteration counter from $h$ to $sh + j$ where $0 \leq h < H'$ is the outer iteration counter (where communication occurs) and $1 \leq j \leq s$ is the inner iteration counter (where a sequence of $s$ solution vectors are computed). By induction we can show that
\begin{align}
x_{sh+j} &= x_{sh} + \sum_{i=1}^{j-1} \frac{\eta_{sh+i}}{m}\tilde A^T\mathbb{I}^T_{sh+i}sig\left(\mathbb{I}_{sh+i}\tilde A x_{sh+i}\right) \nonumber\\  & + \frac{\eta_{sh+j}}{m} \tilde A^T \mathbb{I}_{sh+j}^T sig\biggr( \mathbb{I}_{sh+j}\tilde A x_{sh}\nonumber\\ &+ \sum_{i = 1}^{j-1} \frac{\eta_{sh+i}}{m} \mathbb{I}_{sh+j}\tilde A \tilde A^T\mathbb{I}_{sh + i}^Tsig\left(\mathbb{I}_{sh+i}\tilde A x_{sh+i}\right)\biggr)\label{eq:caupdate}.
\end{align}
Note that we omit the expansion of $x_{sh+i}$ in \eqref{eq:caupdate} for clarity and will show how it is handled in Section \ref{sec:proofs}. The resulting CA-SGD algorithm is shown in Algorithm \ref{alg:casgd}. 
%\begin{figure*}
%\begin{center}
%\includegraphics[trim={0cm 5cm 13cm, 1cm},clip,width=1.5\columnwidth]{matvec}
%\end{center}
%\caption{Visualization of the distributed matrix-vector products $v = Ax$ and $w = A^Tv$. Since $A$ is stored in 1D-block column layout. $v = Ax$ can be computed by multiplying local columns of $A$ (areas within the orange lines) with corresponding entries of $x$. Each processor will now have a partial solution which must be sum-reduced across all processors to obtain $v$. Since each processor has its own copy of $v$, $w = A^Tv$ can be computed by multiplying local rows of $A^T$ with $v$ to get a distributed copy of $w$. In general, vectors in the column-space of $A$ will be replicated and vectors in the row-space will be distributed. \textcolor{red}{\textbf{Omit figure for TPDS? Illustrate Gram matrix computation for 1D-block column and 1D-block row instead.}}}
%\label{fig:matvec}
%\end{figure*}
%\textcolor{red}{Omit if figure is deleted.} Figure \ref{fig:matvec} illustrates the matrix-vector product $\mathbb{I}_h\tilde Ax_{h-1}$ and, then, the matrix-vector product $\tilde A^T\mathbb{I}_h^T$ after the nonlinear operations on the vector $\mathbb{I}_h\tilde Ax_{h-1}$. As shown in the figure, only one all-reduce is required between the two matrix-vector products.
%Might be able to avoid communication with row-partitioned layout too, if the $A^T$ matvec is deferred until the end. I think it will just be 2 reductions for s iterations instead of 1 every iteration. 

%!TEX ROOT=../calogistic-tpds.tex
\section{Analysis of Algorithms}\label{sec:proofs}
Note that CA-SGD requires the matrix-matrix multiplications $\mathbb{I}_{sh+j}\tilde A \tilde A^T\mathbb{I}_{sh+i}^T$ in addition to the matrix-vector multiplications using $\tilde A^T\mathbb{I}_{sh+i}sig(\cdot)$ and $\mathbb{I}_{sh+i}\tilde Ax_{sh+i}$. Unlike prior work on ridge regression, LASSO, and SVM \cite{mythesis,siscpaper,ipdpspaper,soori18}, the nonlinear vector operation $sig(\cdot)$ prevents simplification of \eqref{eq:caupdate}. We will show in this section that CA-SGD can reduce the latency cost despite the nonlinearity under two data partitioning schemes (1D-block column and 1D-block row). Due to the nonlinearity in \eqref{eq:caupdate} we will rely on the communication properties of the distributed matrix-vector products $v = Ax$ and $w = A^Tv$ in order to avoid communication in the $x_h$ update step.% Note that the prior work on linear/ridge regression, LASSO, and SVM  deal with linear or piecewise linear function. This linearity results in further simplification in the CA algorithm's gradient computation step. However, logistic regression requires the non-linear sigmoid function during the gradient computation.
%We will begin by proving lemmas related to the matrix-vector products and use those results to prove running time costs of SGD and CA-SGD. 
\begin{lemma}\label{thm:matvec}
Given a matrix $A \in \mathbb{R}^{m \times n}$ stored in 1D-block column format across $p$ processors and a vector $x \in \mathbb{R}^n$ distributed across the $p$ processors, the matrix vector product $v = Ax$ with $v$ replicated on all processors requires $O\left(m\right)$ words moved and $O\left(\log p \right)$ messages.
\end{lemma} 
\begin{IEEEproof}
Computing $Ax$ requires that each row  of $A$ be multiplied by $x$. With the given partitioning, each processor can multiply the portion of row elements stored locally with the corresponding locally stored elements of $x$. Each processor produces a partial vector $v^{(i)} \in \mathbb{R}^m~\forall~i \in \{1,2,\ldots, p\}~s.t.~v = \sum_{i=1}^p v^{(i)}$. Computing $\sum_{i=1}^p v^{(i)}$ and replicating $v$ on all processors requires one all-reduce with summation which costs $O\left(m\right)$ words moved and $O\left(\log p\right)$ messages. Note that the MPI implementation makes runtime decisions about the optimal routing algorithm based on message-size and number of processors \cite{thakur05}. This proof selects bandwidth and latency bounds with the lowest latency cost. 
\end{IEEEproof}
\begin{lemma}\label{thm:matvec2}
Given a matrix $A \in \mathbb{R}^{m \times n}$ stored in 1D-block column format across $p$ processors and a vector $v \in \mathbb{R}^m$ replicated on all $p$ processors, the matrix vector product $w = A^Tv$ with $w$ distributed across the $p$ processors does not require comunication.
\end{lemma}
\begin{IEEEproof}
A similar cost analysis to Lemma \ref{thm:matvec} proves this lemma.%Since the vector $v$ is replicated on all processors, each processor can compute a portion of $w$ by multiplying locally stored rows of $A^T$ with all of $v$. The resulting vector $w$ is already distributed across processors after the computation, hence no communication is required.
\end{IEEEproof}
%Lemma \ref{thm:matvec2} shows that with $A$ stored in 1D-block column layout the matrix vector product $w = A^Tv$ does not require communication. We will use this fact to show that the CA-SGD method can asymptotically reduce the communication cost of SGD.
From Lemmas \ref{thm:matvec} and \ref{thm:matvec2} we can see that in \eqref{eq:caupdate} computing $\mathbb{I}_{sh+i}\tilde Ax_{sh+i}$ requires communication, whereas apply the sigmoid function and computing $\tilde A^T \mathbb{I}_{sh+j}^Tsig\left(\mathbb{I}_{sh+i} \tilde A x_{sh+i}\right)$ does not.

We will now analyze the computation and communication costs of SGD and CA-SGD. We assume that $A$ is sparse with nonzeros distributed uniformly between the rows and that the vectors $y$ and $x$ are dense. We will use the notation $fmn$ to refer to the number of nonzeros in $A$, where $0 < f \leq 1$. This allows us to bound the number of nonzeros of $\mathbb{I}_h\tilde A$ by $\text{nnz}(\mathbb{I}_h\tilde A) = {fbn}$. Furthermore, we will also assume that the nonzeros are distributed uniformly between the processors. Note that logistic regression requires an $\exp(\cdot)$ operation and element-wise division on $b$-dimensional vectors. This operation requires more floating-point operations to compute than typical arithmetic operations. We will model this by introducing the parameter $\omega$ to represent the cost of a single $sig(\cdot)$ operation. The elementwise $sig(\cdot)$ operation on a $b$-dimensional vector, as a result, costs $\omega b$ flops.

\begin{theorem} \label{thm:lrsgd}
Given a matrix $A \in \mathbb{R}^{m \times n}$ stored in 1D-block column layout on $p$ processors, labels $y \in \mathbb{R}^m$ replicated on all processors, and $x \in \mathbb{R}^n$ partitioned across $p$ processors, $H$ iterations of SGD (Alg. 2) with batch size $b$ requires $O\left(H \frac{fbn}{p} + H\frac{n}{p} + H\omega b\right)$ flops,  $O\left(Hb\right)$ words moved, and $O\left(H\log p\right)$ messages sent.
\end{theorem}
\begin{IEEEproof}
Each iteration of SGD requires computation of $\mathbb{I}_h \tilde A x_{h-1}$ which costs $ \frac{fbn}{p}$ flops and produces a $b$-dimensional vector on each processor which must be sum-reduced. The all-reduce with summation requires $b$ words moved and $\log p$ messages. Computing $sig(\mathbb{I}_h \tilde A x_{h-1})$ costs $\omega b$ flops and no communication. The matrix-vector product $\frac{\eta_h}{m}\tilde A^T\mathbb{I}_h^Tsig\left(\mathbb{I}_h\tilde Ax_{h-1}\right)$ costs $ \frac{fbn}{p}$ multiplications and does not require communication (from Lemma \ref{thm:matvec2}). Finally, updating $x_h$ costs $\frac{n}{p}$ flops and does not require any communication. Multiplying each cost by $H$ gives the results of this proof.
\end{IEEEproof}

%By setting $b = m$ or $b = 1$ we can specialize Theorem \ref{thm:lrsgd} to gradient descent and SGD, respectively. When comparing the two extremes, we can observe that SGD requires less computation than gradient descent at every iteration, but requires $m\over b$ iterations (and rounds of communication) in order to perform one pass over $A$ (i.e. one iteration of GD). This suggests that the computational efficiency of SGD comes at the cost of communication. In a distributed-memory setting, the cost of communicating at every iteration is likely to be the dominant cost. 
We will now show that our new CA-SGD algorithm can asymptotically reduce the communication cost.

\begin{theorem}\label{thm:lrcasgd}
Given a matrix $A \in \mathbb{R}^{m \times n}$ stored in 1D-block column layout on $p$ processors, labels $y \in \mathbb{R}^m$ replicated on all processors, and $x \in \mathbb{R}^n$ partitioned across $p$ processors, $H$ iterations of CA-SGD (Alg. 3) with batch size $b$ requires $O\left(H \frac{f^2sb^2n}{p} + H\frac{n}{p} + Hsb^2 + H\omega b\right)$ flops,  $O\left(Hsb^2\right)$ words moved, and $O\left(\frac{H}{s}\log p\right)$ messages sent.
\end{theorem}
\begin{IEEEproof}
Each iteration of CA-SGD begins by computing the matrix vector product $\begin{bmatrix} \mathbb{I}_h \\ \mathbb{I}_{h+1} \\ \ldots \\ \mathbb{I}_{h+s} \end{bmatrix} \tilde A x_{h-1}$. This computation requires $O\left(\frac{fsbn}{p}\right)$ flops. In addition to the matrix vector product, the Gram matrix $\begin{bmatrix} \mathbb{I}_h \\ \mathbb{I}_{h+1} \\ \ldots \\ \mathbb{I}_{h+s} \end{bmatrix} \tilde A \tilde A^T \begin{bmatrix} \mathbb{I}_h^T & \mathbb{I}_{h+1}^T & \ldots & \mathbb{I}_{h+s}^T \end{bmatrix}$ must be computed. This costs $O\left(\frac{f^2s^2b^2n}{p}\right)$ when computing pair-wise inner products\footnote{Note that the diagonal blocks of the Gram matrix are not required.}. There are $s^2b^2$ possible inner products and each costs $O\left(f^2n\right)$ flops. Once the partial matrix-vector products and Gram matrices are computed, an all-reduce with summation is required to combine the partial products. This communication requires $s^2b^2 + sb$ words moved and $\log p$ messages. Then, we can compute $s$ gradient vectors each of which requires a $b$-dimensional elementwise $sig(\cdot)$ operation. This costs $\omega sb$ flops and no communication. In order to complete the gradient computation, a matrix vector product with blocks of $\tilde A^T$ are required which costs $O\left(\frac{fsbn}{p}\right)$ flops and no communication (from Lemma \ref{thm:matvec2}). Note that after the first gradient is computed, the subsequent $s - 1$ gradients require additional computation in order to correct for missed solution updates. This additional computation requires $s^2b^2$ flops (there are $\frac{(s-2)(s-1)}{2}$ total matrix vector products). Once all gradients have been computed, the solution vector $x_{sh+s}$ can be computed by taking a sum over all $s$ gradients which requires $O\left(\frac{sn}{p}\right)$ flops. Unlike SGD, each iteration of CA-SGD computes $s$ gradients. Therefore, $\frac{H}{s}$ outer iterations of CA-SGD are required to perform the equivalent of $H$ SGD iterations. Multiplying the costs by $\frac{H}{s}$ gives the results of this proof.%$O\left(H\frac{f^2sb^2n}{p} + H\frac{n}{p} + Hsb^2 + H\omega b\right)$ flops, $O\left(Hsb^2\right)$ words moved and $O\left(\frac{H}{s}\log p\right)$ messages.
\end{IEEEproof}
\begin{table*}[t]
\caption{Properties of the LIBSVM Datasets for Numerical Experiments}
\label{tbl:dsetnum}
\centering
\begin{tabular}{|c||c||c||c||c||c||c|}
\hline
Name & $m$ & $n$ & $nnz(A)$ & $nnz(A)/(mn)$ & $\sigma_{\text{max}}$& $\sigma_{\text{min}}> 0$ \\
\hline
a6a & $11,220$ & $123$ & $155,608$& $0.1137$ & $47.9015$ & $0.9972$\\
\hline
Mushrooms & $8,124$ & $112$ & $170,604$ & $0.1875$ & $289.8993$ & $1.2841$\\
\hline
w7a & $24,692$ & $300$ & $288,148$ & $0.0389$  & $9.8112$ & $0.6846$\\
\hline
\end{tabular}
\end{table*}
We will now assume that $A$ is stored in 1D-block row layout, $y$ is distributed across the processors, and $x$ is replicated on all processors. Note that under this partitioning scheme choosing $b$ rows of $A$ uniformly at random may cause load imbalance\footnote{Some processors may have more than the average number of rows chosen from locally stored data while other may have no rows selected.}, so assume that each processor will chose an equal number of local rows (i.e. $b \geq p$ s.t. $b/p \in \mathbb{Z}^+$). This variant of SGD interpolates between sequential SGD when $p = 1$ and GD when $p = m$. Note that this variation has not been discussed in prior work \cite{mythesis, siscpaper,ipdpspaper, soori18}.
\begin{theorem}\label{thm:lrsgd_1drow}
Given a matrix $A \in \mathbb{R}^{m \times n}$ stored in 1D-block row layout on $p$ processors, labels $y \in \mathbb{R}^m$ distributed across all processors, and $x \in \mathbb{R}^n$ replicated on all processors, $H$ iterations of SGD (Alg. 2) with batch size $b$ requires $O\left(H \frac{fbn}{p} + Hn + H \frac{\omega b}{p}\right)$ flops,  $O\left(Hn\right)$ words moved, and $O\left(H\log p\right)$ messages sent.
\end{theorem}
\begin{IEEEproof}
Each iteration of SGD requires computation of $\mathbb{I}_h\tilde A x_{h-1}$ which costs $O\left(\frac{fbn}{p}\right)$ flops since each processor selects $\frac{b}{p}$ rows from locally stored data. Computing $sig(\mathbb{I}_h\tilde A x_{h-1})$ requires each processor to perform the $sig(\cdot)$ operation on a $\frac{b}{p}$-dimensional vector which costs $\frac{\omega b}{p}$ flops. The matrix-vector product $\frac{\eta_h}{m}\tilde A^T\mathbb{I}_h^Tsig(\mathbb{I}_h\tilde Ax_{h-1})$ costs $O\left(\frac{fbn}{p}\right)$ flops and requires an all-reduce with summation. The all-reduce communicates $n$ words using $\log p$ messages. Finally updating $x_h$ costs $n$ flops and no communication since all processors have a copy of $x_{h-1}$ and a copy of the gradient. Multiplying by the number of iterations $H$ provides the results of this proof.
\end{IEEEproof}
\begin{figure*}
\centering
\subfloat[a6a Loss vs. Epochs\label{fig:a6a_objval}]{\includegraphics[width=.29\linewidth]{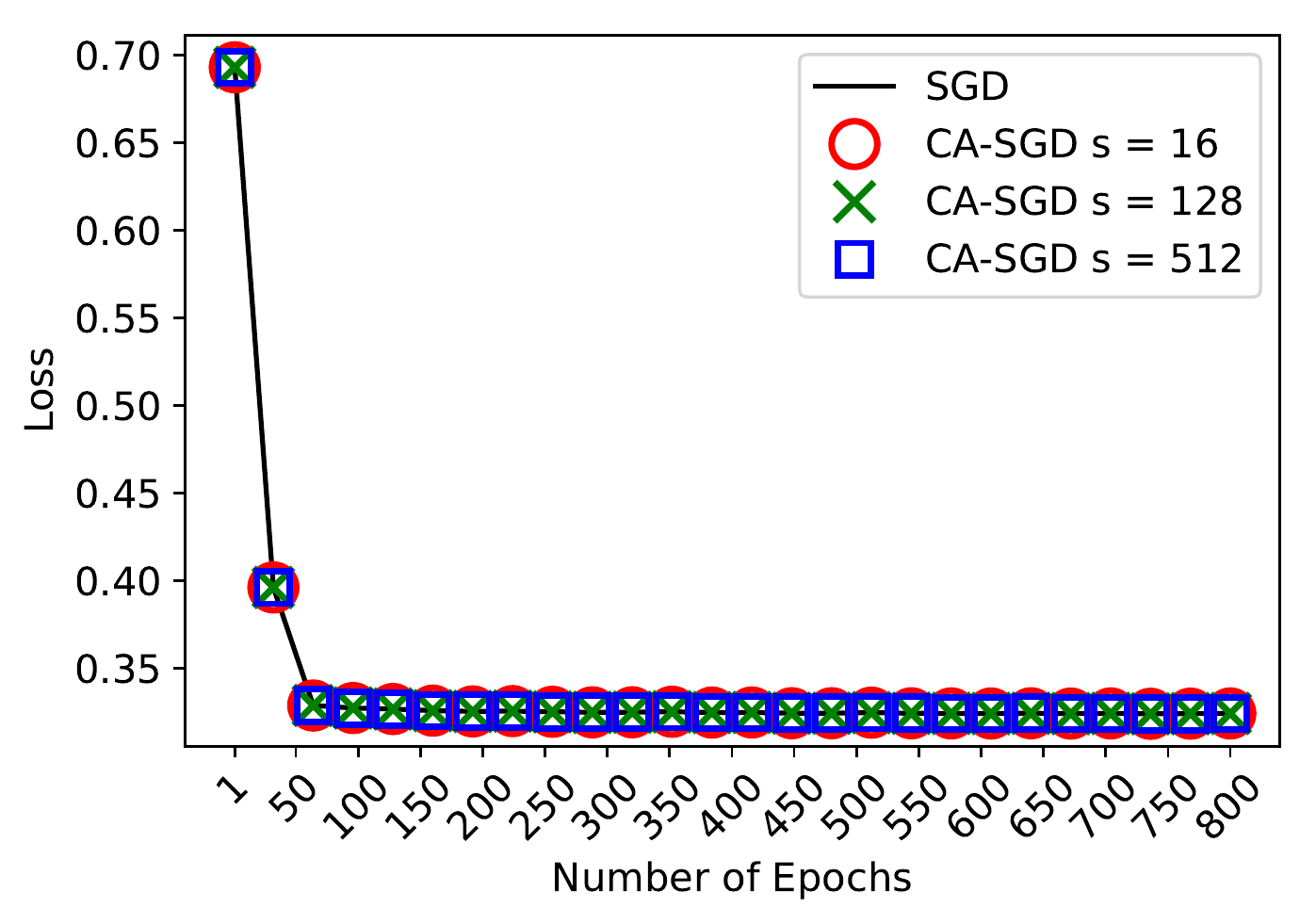}}\qquad
\subfloat[Training Accuracy vs. Epochs\label{fig:a6a_acc}]{\includegraphics[width=.29\linewidth]{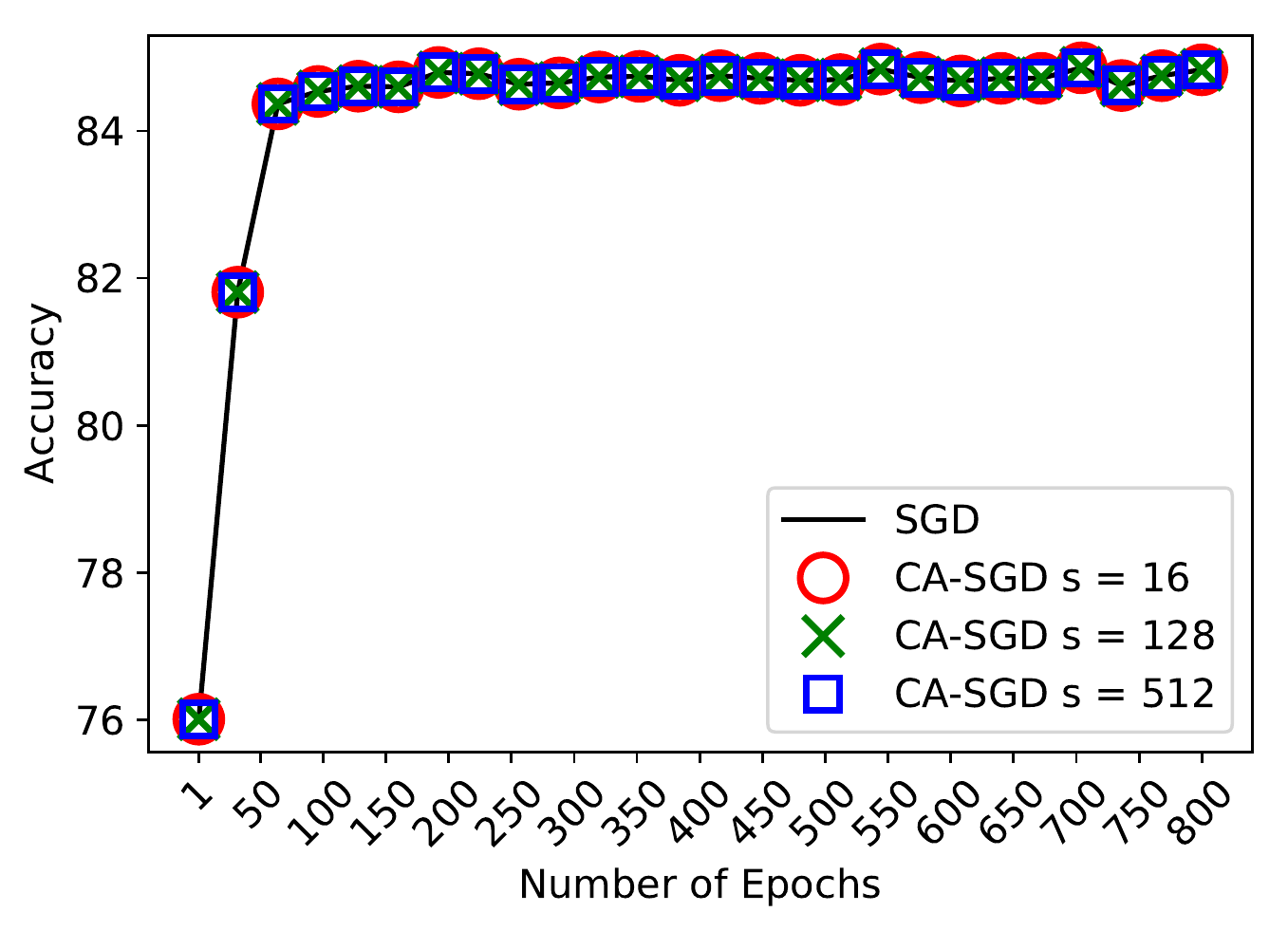}}\qquad
\subfloat[Relative Solution Error vs. Epochs\label{fig:a6a_relerr}]{\includegraphics[width=.29\linewidth]{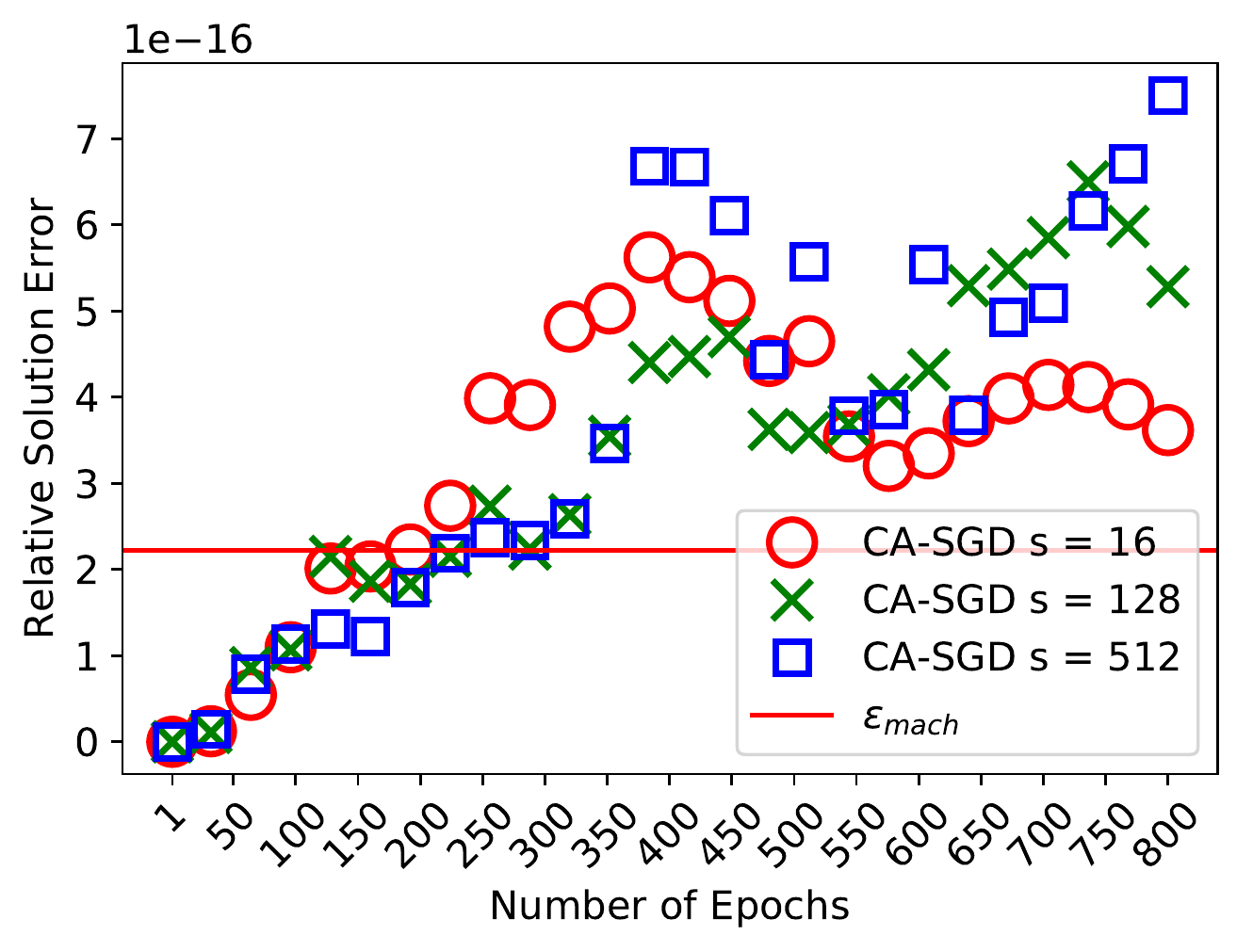}}

\subfloat[Mushrooms Loss vs. Epochs \label{fig:mushrooms_objval}]{\includegraphics[width=.29\linewidth]{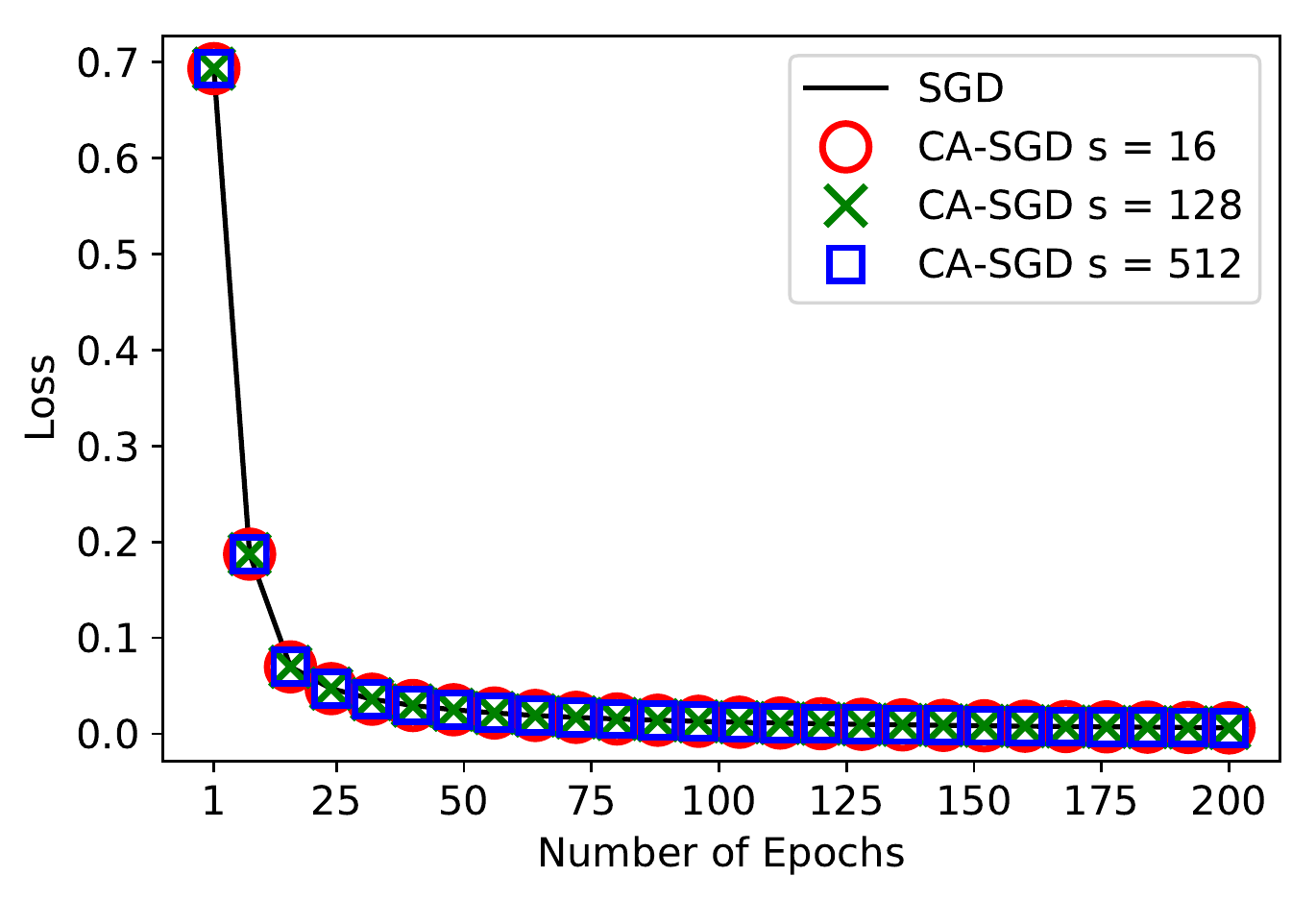}}\qquad
\subfloat[Training Accuracy vs. Epochs\label{fig:mushrooms_acc}]{\includegraphics[width=.29\linewidth]{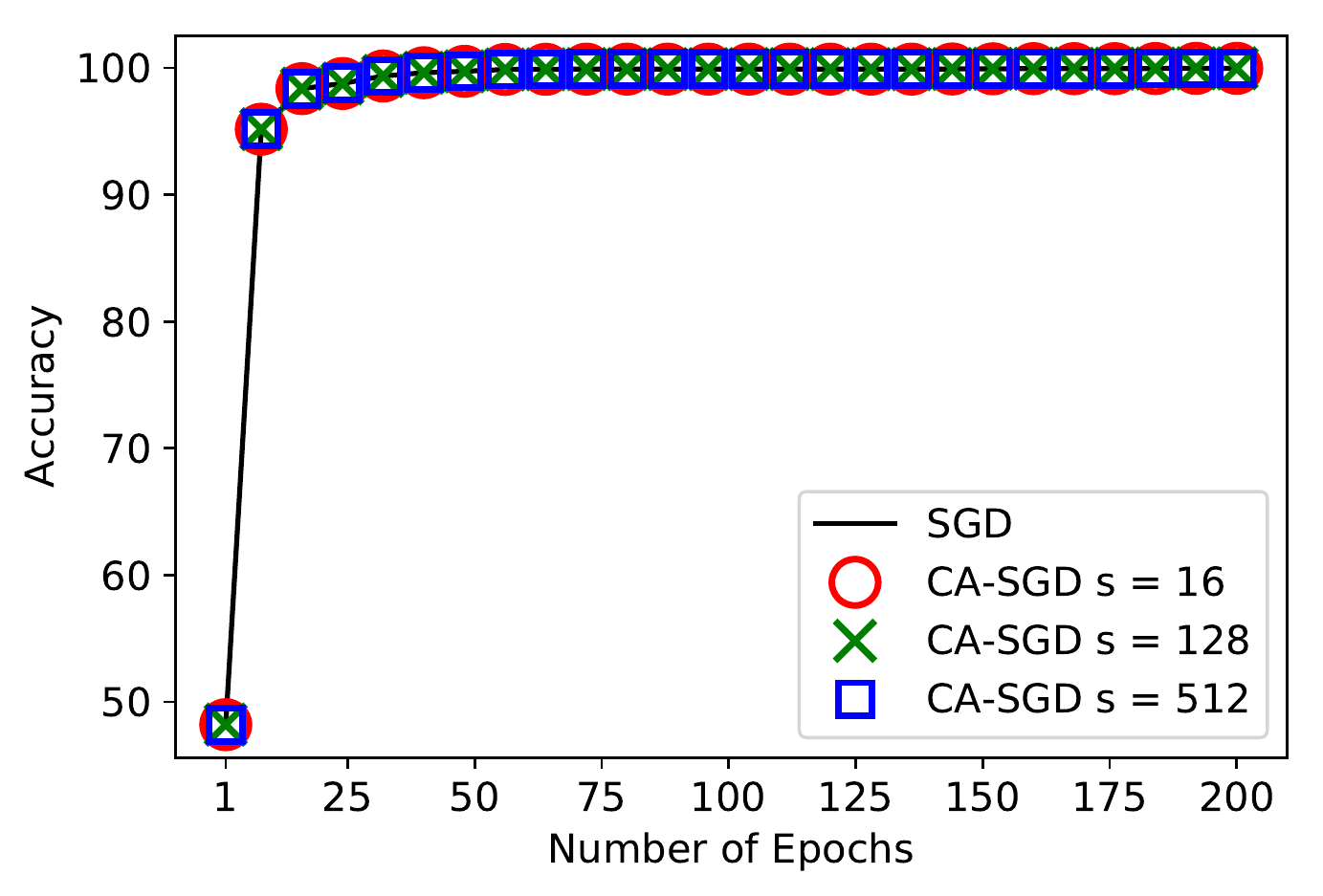}}\qquad
\subfloat[Relative Solution Error vs. Epochs\label{fig:mushrooms_relerr}]{\includegraphics[width=.29\linewidth]{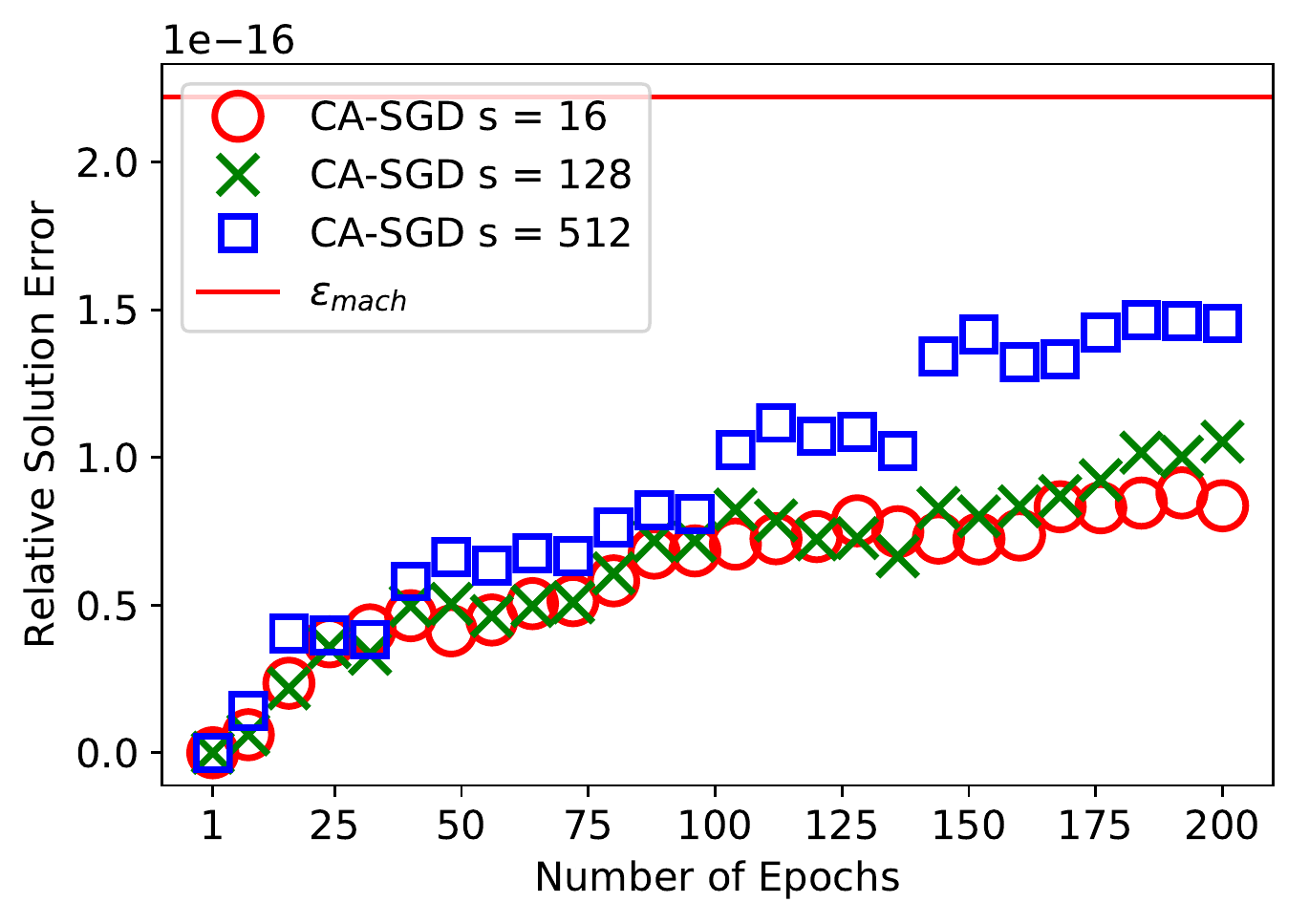}}

\subfloat[w7a Loss vs. Epochs\label{fig:w7a_objval}]{\includegraphics[width=.29\linewidth]{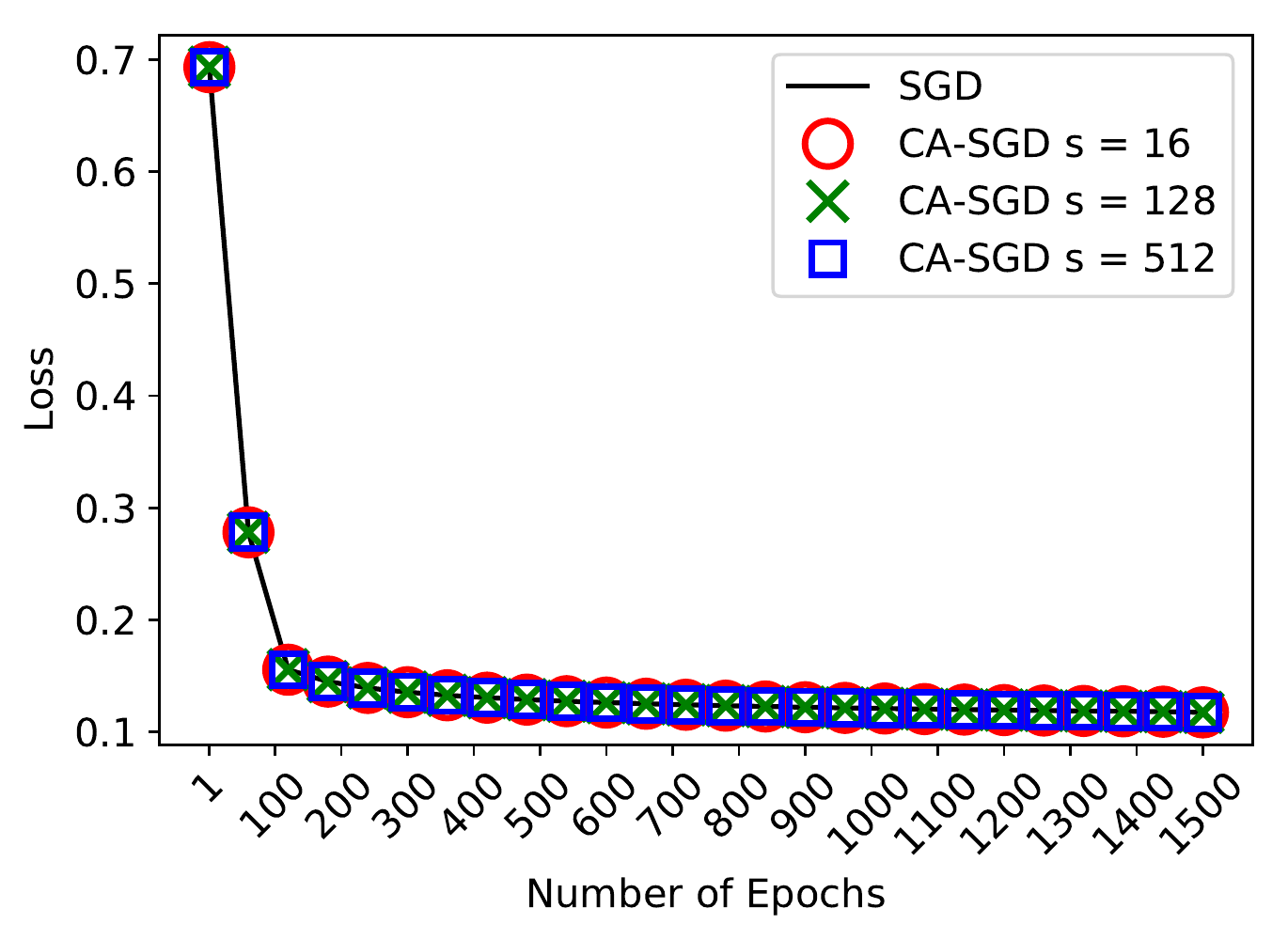}}\qquad
\subfloat[Training Accuracy vs. Epochs\label{fig:w7a_acc}]{\includegraphics[width=.29\linewidth]{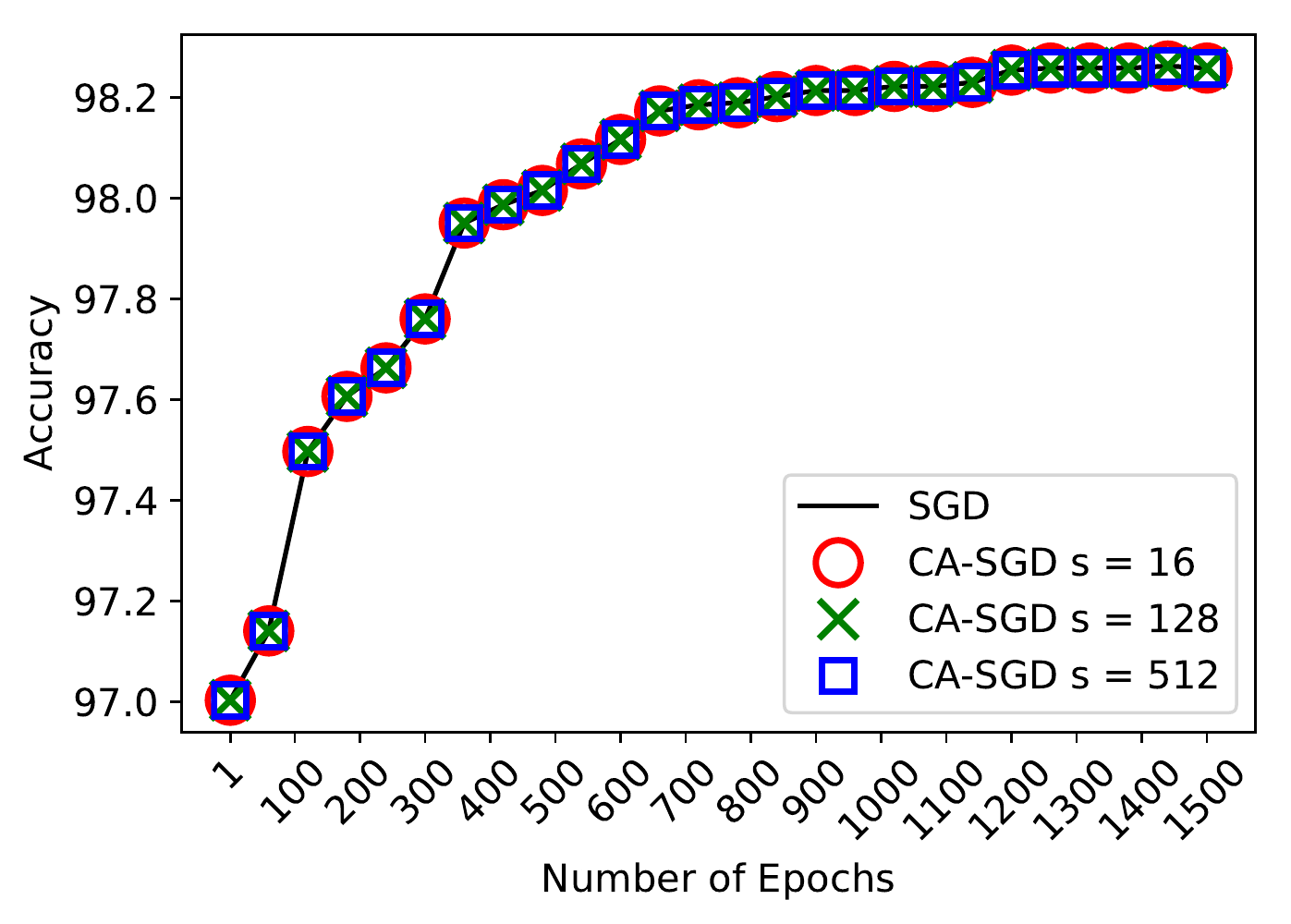}}\qquad
\subfloat[Relative Solution Error vs. Epochs\label{fig:w7a_relerr}]{\includegraphics[width=.29\linewidth]{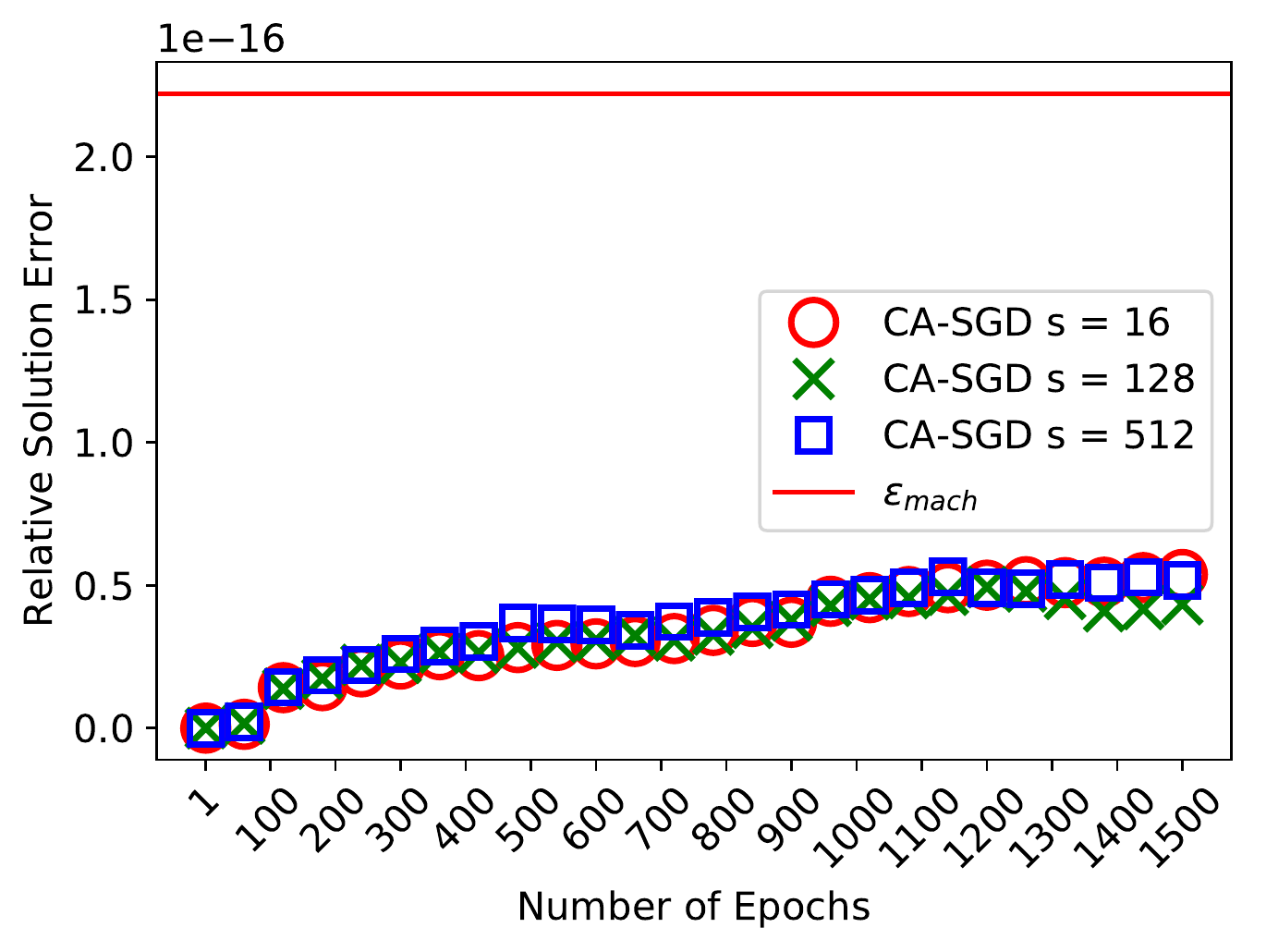}}
\caption{Comparison of SGD and CA-SGD convergence behavior on the a6a, mushrooms, and w7a (see Table \ref{tbl:dsetnum}) for various values of $s$. The loss function, relative solution error, and training accuracy are reported over $100$ epochs of training.}\label{fig:numexp}
\end{figure*}
From this proof we can observe that SGD with 1D-block row layout also requires one round of communication for each iteration. We will now prove the computation and communication cost of CA-SGD with 1D-block row layout.
\begin{theorem}\label{thm:lrcasgd_1drow}
Given a matrix $A \in \mathbb{R}^{m\times n}$ stored in 1D-block row layout on $p$ processors, labels $y \in \mathbb{R}^m$ distributed across all processors, and $x \in \mathbb{R}^n$ replicated on all processors, $H$ iterations of CA-SGD (Alg. 3) with batch size $b$ requires $O(H\frac{f^2sb^2n}{p} + Hn + H\frac{sb^2}{p} + H\frac{\omega b}{p})$ flops, $O(Hfbn + Hb + H\frac{n}{s})$ words moved, and $O(\frac{H}{s}\log p)$ messages sent.
\end{theorem}
\begin{IEEEproof}
Each iteration of CA-SGD requires computation of $Yx_{sh}$ where $Y = \begin{bmatrix}\mathbb{I}_{sh+1}\\ \mathbb{I}_{sh+2}\\ \ldots \\ \mathbb{I}_{sh+s}\end{bmatrix}\tilde A $ which costs $O\left(\frac{fsbn}{p}\right)$ flops and no communication. The resulting $sb$-dimensional vector is partitioned across $p$ processors. We must also compute the Gram matrix $G = YY^T$. Since $Y$ is 1D-block row partitioned across $p$ processors with each processor storing $\frac{sb}{p}$ rows, computing $G$ requires communication. By using an all-gather routine \cite{thakur05}, each processor can obtain the necessary rows of $Y$ from other processors in order to compute $G$. In addition to communicating $Y$ we also communicate the vector $Yx_{sh}$. This costs $O\left(\frac{f^2s^2b^2n}{p}\right)$ flops and communicates $O\left((s-1)fbn + sb\right)$ words using $\log p$ messages\footnote{If the message size for the all-gather is large, then many MPI implementations will switch to a 1D-ring routing algorithm. CA-SGD reduces the latency cost by a factor of $s$ in both situations.}. Each processor computes blocks of $G$ such that $G$ is stored in 1D-block cyclic row layout with each processor storing $\frac{b}{p}$ consecutive rows in each $b \times b$ upper triangular block of $G$). We can now compute all $s$ of the $sig(\cdot)$ vectors shown in \eqref{eq:caupdate} which requires $\frac{s^2b^2 + \omega sb}{p}$ flops on each processor (the resulting $b$-dimensional vectors are partitioned across all processors). We can now compute the gradient by multiplying the local $\frac{sb}{p}$ columns of $\tilde A^T$ with the local $\frac{sb}{p}$ elements of the $sig(\cdot)$ vectors. This costs $\frac{fsbn}{p}$ flops and no communication. The resulting $s$ gradient vectors on each processor are $n$-dimensional and must be sum-reduced in order to update the solution vector. The all-reduce communicates $n$ words\footnote{Note that we perform local summations on the $s$ gradient vectors to obtain one partially summed vector that is subsequently sum-reduced across processors. However, doing so means we can only obtain the final solution vector $x_{sh+s}$ and not the intermediate solutions $x_{sh+j}~\forall~j=1,2,\ldots,s-1$.} using $\log p$ messages. Once all processors have a copy of the $s$ gradient vectors, they can be summed to obtain the update to the solution vector, which costs $sn$ flops. Combining the costs results in $O\left(\frac{f^2s^2b^2n}{p} + sn + \frac{s^2b^2}{p} + \frac{\omega sb}{p}\right)$ flops and communicates $O\left(fsbn+sb + n\right)$ words in $O\left(\log p\right)$ messages per outer iteration. Multiplying the per iteration costs by $\frac{H}{s}$ gives the results of this proof.
\end{IEEEproof}
These proofs show that CA-SGD reduces the latency cost by a tunable factor of $s$ at the expense of additional bandwidth and computation cost. If latency is the dominant cost, then CA-SGD can attain $s$-fold speedup over SGD. This section primarily focuses on 1D layouts of $A$, however, 2D layouts of $A$ might yield better performance for large, (nearly) square matrices. In this setting we can combine a divide-and-conquer local SGD algorithm with our (1D-column layout) CA-SGD method. %This variant would have each row team of processors perform a tunable number of CA-SGD iterations (parallelized over the column team of processors) on locally assigned rows of $A$. The resulting solution vectors from each row team can then be sum-reduced or averaged across rows in order to obtain a global solution vector. Note that the divide-and-conquer approach (across the row team of processors) is known to converge from the CoCoA/prox-CoCoA work \cite{cocoa, forte15, ma17, jaggi15}. Furthermore, asynchronously averaging the solutions across rows would yield a distributed-memory variant of HOGWILD! \cite{hogwild}.
We leave the analysis, implementation, and performance comparison of this 2D layout variant for future work.
%\begin{figure*}[t!]
%\includegraphics[width=\columnwidth]{../figures/GD_SGD_objval}
%\includegraphics[width=\columnwidth]{../figures/GD_SGD_accuracy}
%\end{figure*}

%!TEX root=../calogistic-tpds.tex
\section{Experimental Results}
Prior work on CA-Krylov methods \cite{carson,hoemmen,Carson13,Carson14,Carson15} illustrated that applying the CA-technique can result in numerical instability due to the additional computation and rearragement of the solution updates. This held true for even modest values of $s$ and required development of residual replacement and orthogonal basis functions. However, unlike prior work, we will show that CA-SGD is numerically stable for very large values of $s$. Then we will show the practical performance tradeoffs and scaling properties of CA-SGD with an MPI implementation targeting a high-performance Infiniband cluster.
\subsection{Numerical Experiments}
We will now show how the convergence behavior of SGD compares to CA-SGD as $s$ is varied. The datasets used in the experiments are binary classification problems obtained from the LIBSVM repository \cite{libsvm}. Table \ref{tbl:dsetnum} summarizes properties of the datasets tested in this section. The SGD and CA-SGD methods have been implemented in Python using NumPy for linear algebra subroutines.
\begin{figure*}[t]
\centering
\subfloat[news20 scaling\label{fig:news20_scaling}]{\includegraphics[width=.3\linewidth]{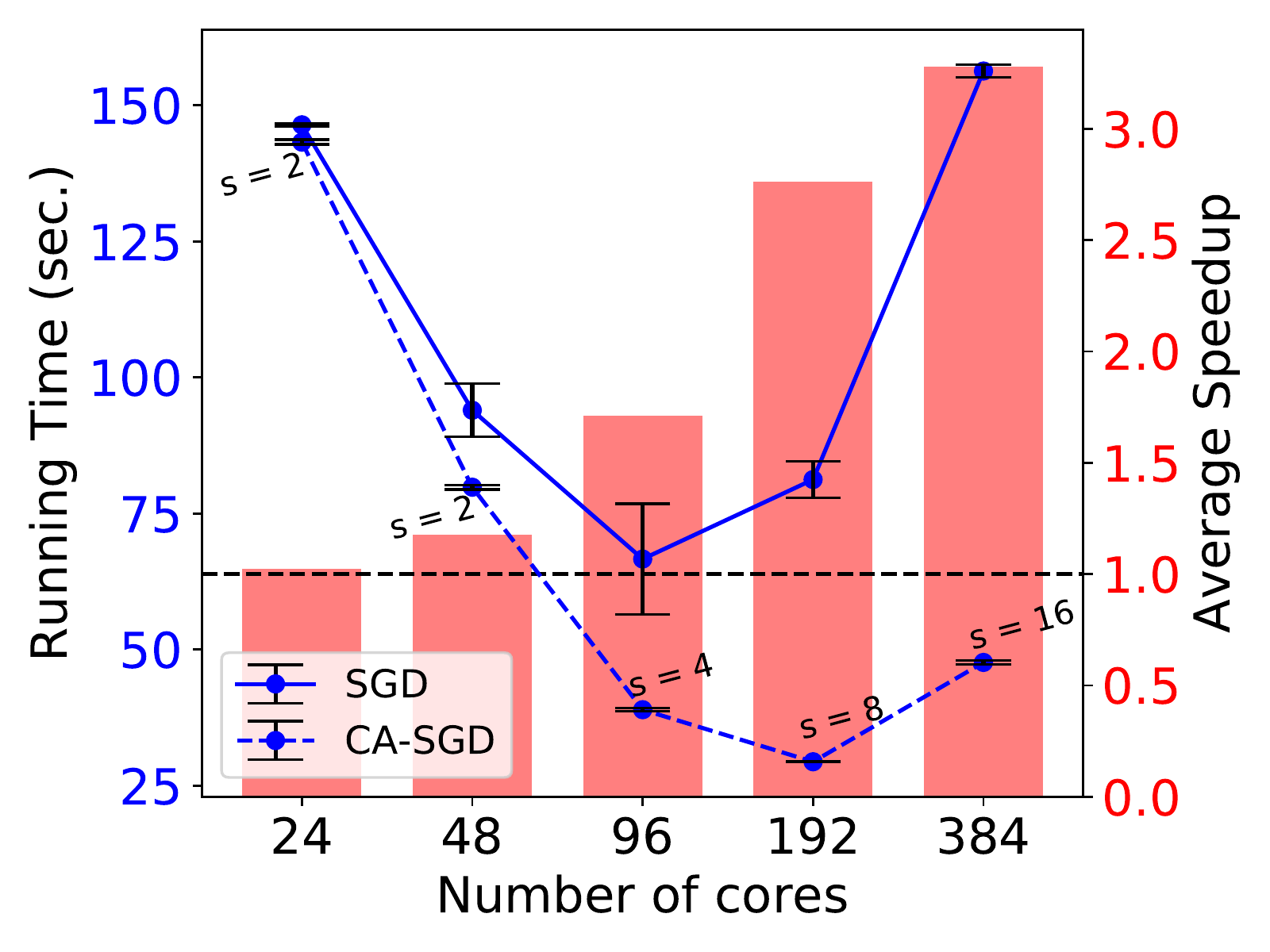}}\qquad
\subfloat[real-sim scaling\label{fig:real-sim_scaling}]{\includegraphics[width=.3\linewidth]{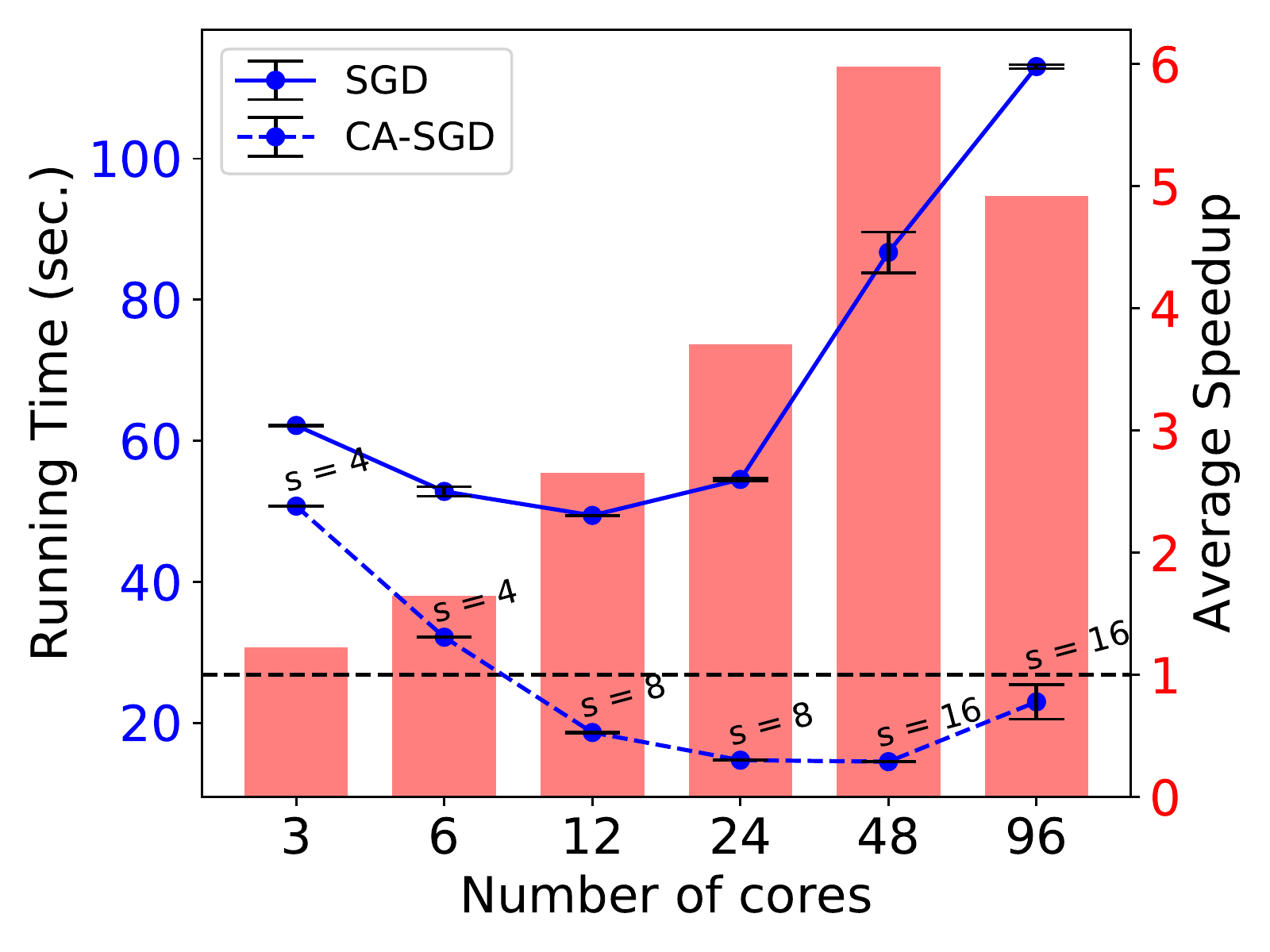}}\qquad
\subfloat[url scaling\label{fig:url_scaling}]{\includegraphics[width=.3\linewidth]{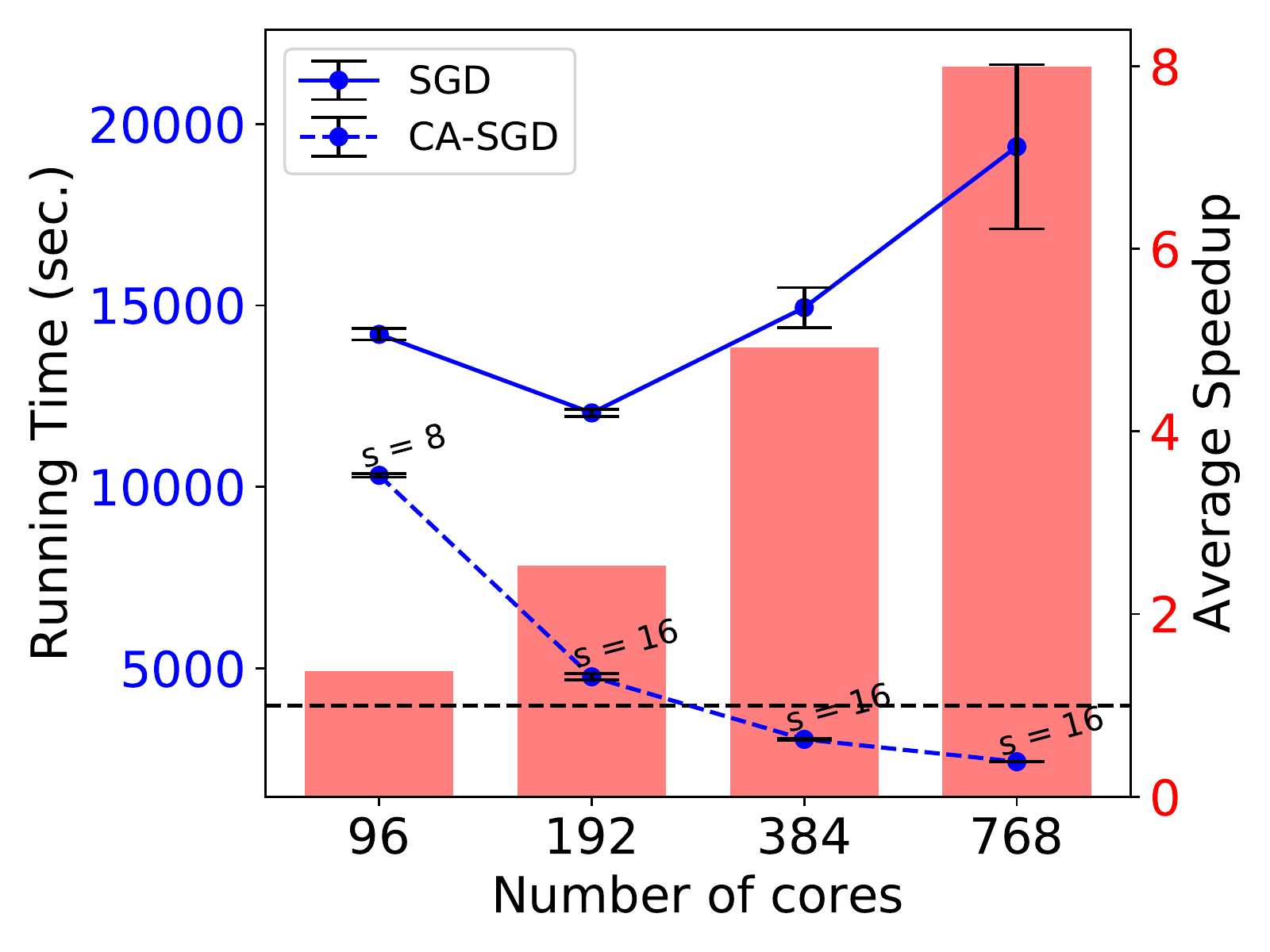}}
\caption{Strong scaling comparison between SGD ($b = 1$) and CA-SGD on the news20, real-sim and url datasets (see Table \ref{tbl:dsetperf}). We report the mean running times (blue points) and the standard deviation (error bars) over 5 trials. CA-SGD results are annotated with the value of $s$ that achieved the fastest running times.}\label{fig:strongscaling}
\end{figure*}
Figure \ref{fig:numexp} illustrates the loss function convergence, relative solution error, and training accuracy of SGD and CA-SGD. From Figures \ref{fig:a6a_objval}, \ref{fig:mushrooms_objval}, and \ref{fig:w7a_objval} we can observe that CA-SGD converges at the same rate as SGD and to the same final loss value for all datasets and values of $s$ up to $512$. Similarly, Figures \ref{fig:a6a_acc}, \ref{fig:mushrooms_acc}, and \ref{fig:w7a_acc} show that CA-SGD attains the same training accuracy as SGD for all values of $s$. The loss convergence and training accuracy figures experimentally validate that CA-SGD is simply a mathematical reformulation and computes the same sequence of partial solutions (up to floating-point error) as SGD.

The next set of experiments aim to quantify the floating-point error in the partial solutions computed by CA-SGD when compared to SGD. To show this, we will plot the relative solution error of CA-SGD with respect to SGD. For these experiments relative solution error is defined as, $\left\lVert x_h - x'_h\right\rVert_2/\left\lVert x_h\right\rVert_2$, where $h$ is the epoch of training, $x_h$ is the SGD solution vector, and $x'_h$ is the CA-SGD solution vector. We will also plot the machine precision, $\epsilon_{\text{mach}}$, of the target computer as a reference line. Figures \ref{fig:a6a_relerr}, \ref{fig:mushrooms_relerr}, and \ref{fig:w7a_relerr} illustrate the results of this experiment. For the mushrooms and w7a datasets, we observe that the relative solution error is below machine precision over all epochs of training, which means there is negligible accumulation of floating-point error from CA-SGD for all values of $s$ that were tested. The relative solution error for the a6a dataset is greater than machine precision but only by a constant factor and is still accurate up to 15-digits. For all datasets we observe that as $s$ increases, the relative solution error also increases. This is to be expected since large values of $s$ require computation of larger Gram matrices and additional matrix-vector products. However, despite the additional computation we see that CA-SGD is numerically stable. 

For other datasets that have similar singular value spread, CA-SGD is likely to remain numerically stable. Furthermore, if techniques like data normalization and regularization are incorporated into the logistic regression model, the datasets become more well-conditioned. As a result, we do not expect CA-SGD to exhibit numerical instability for most practical applications and desired accuracies. In addition, numerical analysis of CA-SGD would be helpful in provide bounds on error accumulation.
%Note that CA-Krylov methods were numerically unstable and required additional techniques (like alternate orthogonal bases and residual replacement strategies) to be introduced in order to correct the instability. In contrast, CA-SGD is numerically stable for all values of $s$ that were tested. However, numerical analysis of CA-SGD would provide insight into how the stability depends on $s$, $b$, and conditioning of $A$.
\subsection{Performance Experiments}
%CA-SGD requires additional computation and bandwidth in order to decrease latency by a factor $s$.
To show the practical tradeoffs between SGD and CA-SGD, we implement both algorithms in C++ with MPI for parallel processing. The input matrix is stored in CSR 3-array format and partitioned in 1D-block column layout. Since SGD and CA-SGD only operate on $b$ and $sb$ rows of $A$, respectively, we reimplemented a subset of sparse BLAS-1 and BLAS-2 so that they operate only on the sampled rows. This avoids the overhead of explicitly copying sampled rows of $A$ into a buffer at every iteration. In addition, we implement a new sparse BLAS-3 kernel to compute the Gram matrix required by CA-SGD. Since the Gram matrix is symmetric, we only compute and store the upper-triangular portion.% Finally, we performed baseline comparisons between random sampling and cyclic sampling of rows of $A$ and found that cyclic sampling yielded faster running times due to better cache efficiency without slowdown in convergence. Therefore, we use cyclic sampling for all subsequent performance experiments. 

The experiments were performed on a high-performance cluster provided by the Maryland Advanced Research Computing Center. The compute nodes are dual-socket 2.5GHz Intel Haswell 12-core processors (24 cores per node) which are interconnected by an Infiniband network using a fat-tree topology. The code is built with the Intel 18.0.3 C++ compiler and OpenMPI 3.1 \cite{openmpi}. We experimented with hybrid OpenMP and MPI configurations but found that flat MPI performed best. Table \ref{tbl:dsetperf} summarizes the LIBSVM \cite{libsvm} datasets used for the performance experiments. All matrices and vectors are stored in double-precision format. The datasets were chosen to illustrate the SGD vs CA-SGD methods at various machine scales (real-sim being small scale, news20 being medium scale, and url being large scale).
\begin{table}[t]
\caption{Properties of the LIBSVM Datasets for Performance Experiments}
\label{tbl:dsetperf}
\centering
\begin{tabular}{|c||c||c||c|}
\hline
Name & $m$ & $n$ & $nnz(A)$\\
\hline
news20 & $19,996$ & $1,355,191$ & $1,674,113$\\
\hline
real-sim & $72,309$ & $20,958$ & $3,709,083$\\
\hline
url & $2,396,130$ & $3,231,961$ & $277,058,644$ \\
\hline
\end{tabular}
\end{table}
\subsubsection{Scaling}
This benchmark is intended to show how CA-SGD and SGD behave as the number of cores is varied. As the number of cores increase, the SGD running time becomes more latency dominant. Since CA-SGD reduces latency cost by $s$, we expect to see performance improvements over SGD. Figures \ref{fig:news20_scaling}-\ref{fig:url_scaling} illustrate the strong scaling (left y-axis in blue) and speedups (right y-axis in red) for the datasets in Table \ref{tbl:dsetperf}. Each plot in Figure \ref{fig:strongscaling} shows the SGD (solid blue) and CA-SGD (dashed blue) running times with batch size of $1$ for each dataset. Both methods were trained for 100 epochs and we report the mean running time and standard deviation (error bars) over 5 trials. The CA-SGD running times are annotated with the value of $s$ that achieved the best performance. 

In Figure \ref{fig:news20_scaling} we can observe that at small scale ($p = 24$ and $p = 48$) the computational cost dominates with $s = 2$ resulting in the best CA-SGD running times. Since the latency cost increases with the number of cores, we can see that CA-SGD can use larger values of $s$ at larger core counts. This eventually leads CA-SGD ($p = 192$ and $s = 8$) to attain an average speedup of $2.27\times$ over SGD ($p = 96$). Finally, at $p = 384$ we see that CA-SGD performance degrades despite increasing $s$. This is because the additional computation and bandwidth costs dominate the reduction in latency cost. Figure \ref{fig:real-sim_scaling} shows the scaling results for the smaller real-sim dataset. This dataset has fewer columns per core which means that the latency cost is more dominant at smaller scales. This is evidenced by the fact that we can start at $s = 4$ for this dataset. As the number of cores increases, we observe that $s$ becomes larger and the speedup from CA-SGD increases. However, at $p = 96$ we see the additional computation and bandwidth costs begin to dominate and performance of CA-SGD degrades. For the real-sim dataset, CA-SGD ($p = 48$ and $s = 16$) achieves an average speedup of $3.41\times$ over SGD ($p = 12$). Figure \ref{fig:url_scaling} illustrates scaling results for the larger url dataset. Due to the size of this dataset, SGD and CA-SGD can scale to larger numbers of cores (potentially higher latency costs). For this dataset, CA-SGD ($p=768$ and $s = 16$) achieves an average speedup of $4.97\times$ over SGD ($p = 192$) and scales to $4\times$ as many cores.
%\begin{figure*}
%\subfloat[news20 loss\label{fig:news20_loss}]{\includegraphics[width=.3\linewidth]{figures/news20_loss.pdf}}\qquad
%\subfloat[real-sim loss\label{fig:real-sim_loss}]{\includegraphics[width=.3\linewidth]{figures/real-sim_loss.pdf}}\qquad
%\subfloat[url loss\label{fig:url_loss}]{\includegraphics[width=.3\linewidth]{figures/url_loss.pdf}}
%
%
%\subfloat[news20 accuracy\label{fig:news20_accuracy}]{\includegraphics[width=.3\linewidth]{figures/news20_accuracy.pdf}}\qquad
%\subfloat[real-sim accuracy\label{fig:real-sim_accuracy}]{\includegraphics[width=.3\linewidth]{figures/real-sim_accuracy.pdf}}\qquad
%\subfloat[url accuracy\label{fig:url_accuracy}]{\includegraphics[width=.3\linewidth]{figures/url_accuracy.pdf}}
%\caption{Comparison of the SGD vs CA-SGD loss and accuracy behavior (with $b = 1$) for various values of $s$ on the news20, real-sim, and url datasets. A reference line (red) is added to show the final loss and accuracy for each dataset after 100 epochs of training. Loss and Accuracy behavior are reported for select values of $s$ for CA-SGD in order to highlight values of $s$ where performance degrades due to additonal bandwidth and computation costs.}
%\end{figure*}
The scaling results in this section suggest that CA-SGD can achieve large average speedups of up to $4.97\times$ over SGD and scale out further on a parallel cluster. These experiments further validate the theoretical analysis and illustrate that reducing latency at the expense of bandwidth and computation can lead to significant performance improvements.
\begin{figure*}
\centering
\subfloat[news20 Running Time Breakdown (p = 48)\label{fig:news20_p48_breakdown}]{\includegraphics[width=.45\linewidth]{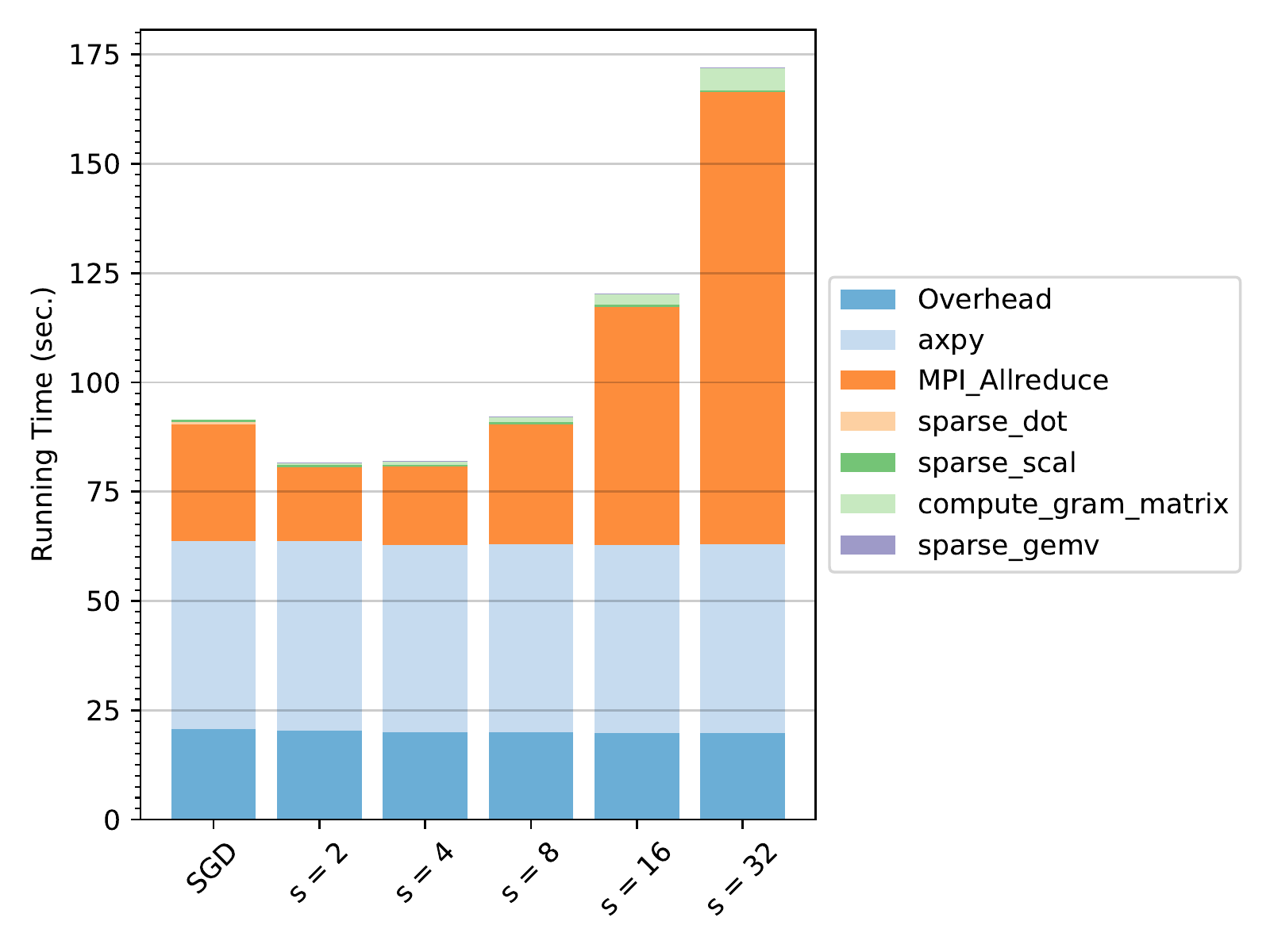}}\qquad
\subfloat[news20 Running Time Breakdown (p = 192)\label{fig:news20_p192_breakdown}]{\includegraphics[width=.45\linewidth]{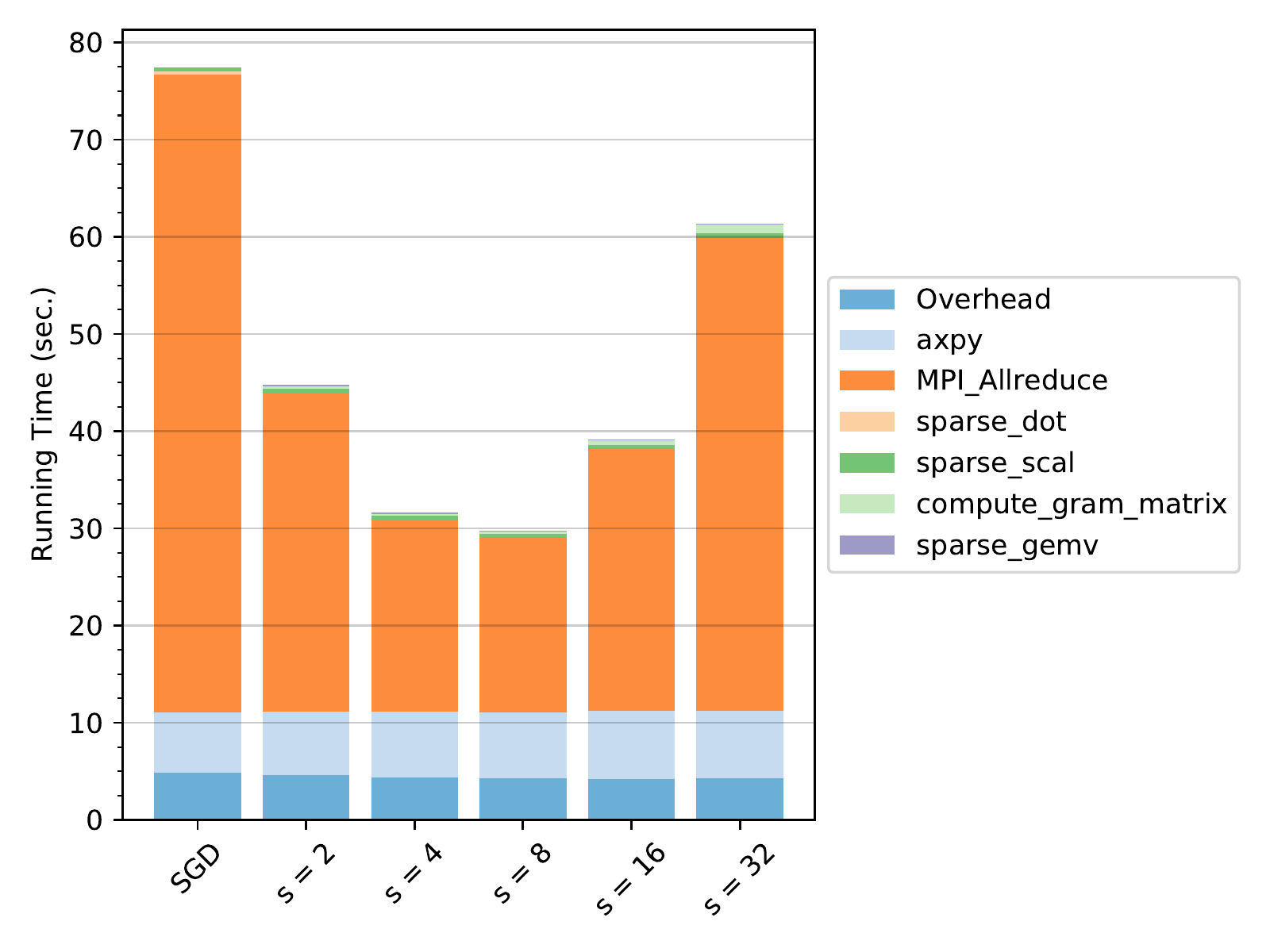}}
\caption{Running time breakdown of SGD and CA-SGD (with $b = 1$) on the news20 dataset (see Table \ref{tbl:dsetperf}) for $p = 48$ and $p = 192$. Overhead includes the time spent on row sampling, scalar operations, loop overhead, and memory management. SGD does not require the sparse gemv nor the compute Gram matrix operations. CA-SGD replaces the sparse dot with sparse gemv computations. Note that MPI\_Allreduce in the legend is a combination of the bandwidth and latency costs analyzed and the remaining legend items except Overhead correspond to the flops cost analyzed in Section \ref{sec:proofs}.}
\end{figure*}
\subsubsection{Running time breakdown}
This benchmark is intended to show a breakdown of how much time is spend on computational kernels and communication routines in the SGD and CA-SGD algorithms. We will compare SGD vs. CA-SGD and at two different core counts. At smaller core counts, latency is less dominant so the benefits of CA-SGD will be less pronounced. However, once we transition to large core counts, the latency reduction of CA-SGD should result in larger speedups. We report the running time breakdown of SGD vs CA-SGD with $b = 1$ for several values of $s$ on the news20 dataset. We obtained the running time breakdown by using the Tuning and Analysis Utilities (TAU) to instrument our code \cite{tau}. %By default TAU throttles functions which take less than $10 \mu s$ and are called $10^6$ times. We disable this throttling feature because SGD and CA-SGD make many calls to BLAS-1 and BLAS-2 operations. In addition, we enable TAU runtime compensation which corrects for instrumentation overhead. 
Since TAU generates profiles for each MPI process, we show the average over all MPI processes. Some operations such as the row sampling, scalar operations, loop overheads, and memory management are grouped into overhead.

Figures \ref{fig:news20_p48_breakdown} and \ref{fig:news20_p192_breakdown} illustrate the running time breakdown of SGD and CA-SGD at $p = 48$ and $p = 192$, respectively. Note that the MPI\_Allreduce times include bandwidth and latency costs. Compute Gram Matrix, sparse dot, sparse scal, sparse gemv, and axpy correspond to the flops cost analyzed in Section \ref{sec:proofs}. At small scale (Fig. \ref{fig:news20_p48_breakdown}), the computation cost (axpy) dominates the communication cost (MPI\_Allreduce). As $s$ increases we begin to see a reduction in MPI\_Allreduce times due to a reduction in latency cost. However, starting at $s = 8$ the additional computation and bandwidth costs of CA-SGD dominate and cause the running times to grow. In the best case (at $s = 2$) CA-SGD achieves a communication speedup of $1.58\times$ and an overall speedup of $1.12\times$ over SGD. 

At large scale (Fig. \ref{fig:news20_p192_breakdown}), the communication time is dominated by latency due to synchronization with $4\times$ more cores. Since latency dominates, CA-SGD achieves speedups for a wider range of values for $s$. At $s = 8$ CA-SGD attains a communication speedup of $6.5\times$ and overall speedup of $2.6\times$ over SGD. In both figures we see cases where CA-SGD is much slower than SGD. In those cases, the additional bandwidth cost of CA-SGD is the bottleneck and not the additional computation. This suggests that if a candidate parallel cluster is bandwidth-limited, then the maximum values of $s$ and the speedups attained by CA-SGD will be limited.
\begin{figure}
\centering
\includegraphics[width=.7\linewidth]{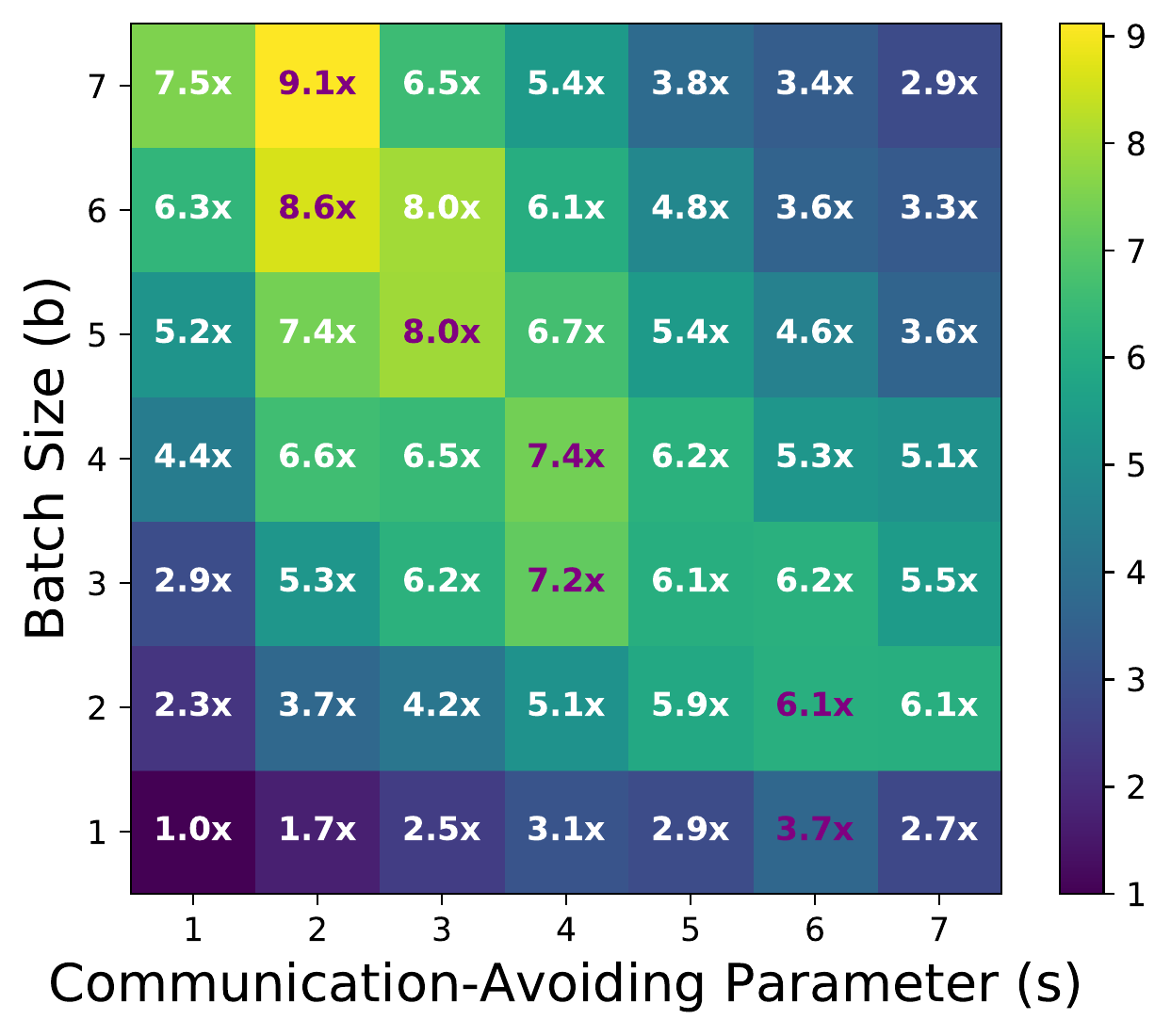}
\caption{Comparison of SGD vs CA-SGD for various batch sizes and values of $s$ on the url dataset with $p = 384$. All speedups are relative to SGD ($ s = 1$ and $b = 1$). As $s$ and $b$ increase, the bandwidth cost for CA-SGD increases by a factor of $sb$. Speedups relative to SGD with $b > 1$ can be obtained by dividing CA-SGD ($s > 1$, $b > 1$) speedups by the corresponding SGD ($s = 1$, $b > 1$) speedup.}
\label{fig:url_heatmap}
\end{figure}
\subsubsection{Batch size vs $s$}
This benchmark will explore how setting $b > 1$ affects the speedups CA-SGD can obtain over SGD. From the previous results with $b = 1$ we see that the additional bandwidth and computation cost introduced by $s > 1$ does degrade CA-SGD performance when $s$ is too high (e.g. $s > 16$ for the url dataset). The performance results thus far compare SGD, which samples a single row every iteration, to its CA-SGD variant. As expected, SGD is latency dominated which results in large speedups for CA-SGD. If the batch size is increased then the $sb$ additional bandwidth and computation costs will likely dominate. %In this section, we will explore how setting $b > 1$ affects the speedups CA-SGD can obtain over SGD. From the previous results with $b = 1$ we see that the additional bandwidth and computation cost introduced by $s > 1$ does degrade CA-SGD performance when $s$ is too high ($s > 16$ for the url dataset). 
We should expect that $s$ must be decreased in order to compensate for increasing the batch size, $b$. Figure \ref{fig:url_heatmap} illustrates a speedup heatmap comparing SGD (with $s = 1$ and $b \geq 1$, bottom-left corner) and CA-SGD (with $s > 1$ and $b \geq 1$) on the url dataset with $p = 384$. Speedups are relative to SGD with $s = 1, b = 1$ (bottom left corner) which means that speedups greater than $1\times$ in column $s = 1$ are due exclusively to increasing batch size. Speedups greater than $1\times$ in row $b =  1$ are due exclusively to communication-avoidance. For data points with $s > 1$ and $b > 1$, speedups greater than $1\times$ are due to a combination of larger batch sizes and communication-avoidance. Note that for $b > 1$, the speedups are due to using BLAS-2 (with increasing matrix sizes) instead of BLAS-1 functions. The heatmap illustrates that CA-SGD is most effective for small batch sizes where latency dominates. %Note that for the url dataset we did not observe any slowdowns in convergence due to increasing the batch size. Increasing the batch size simply resulted in larger speedups. However, in general, speed of convergence is not guaranteed to be the same when $b$ increases.

In these experiments, we focused on square and rectangular sparse matrices stored in CSR format. However, in some situations the input data may be dense. In the dense case, CA-SGD becomes more compute-bound due to the additional elements present in the matrix. As a result, CA-SGD speedups over SGD will be more modest than for sparse matrices. Given that CA-SGD achieves better speedups for latency-dominated/distributed environments, it is well placed to attain large speedups over SGD in cloud environments and when using programming models like Spark/MapReduce (due to the higher latencies in those settings). CA-SGD is unlikely to attain large speedups in shared-memory environments where inter-core latencies are orders of magnitude lower than inter-node latencies. However, exploring and quantifying the performance difference between CA-SGD and SGD on shared-memory would be interesting. We intend to study the performance evaluation of CA-SGD on the various hardware and programming environments in future work.

%!TEX ROOT=../calogistic-tpds.tex
\section{Conclusion and Future Work}
In this paper, we derived a communication-avoiding variant of SGD for solving the logistic regression problem. We proved theoretical bounds on the computation and communication costs which showed that CA-SGD reduces latency costs by a tunable factors of $s$. We showed that CA-SGD is numerically stable and achieves speedups of up to $4.97\times$ over SGD on a high-performance Infiniband cluster. When latency is the dominant cost CA-SGD can achieve large speedups despite the additional bandwidth and computation costs. However, as the computation and bandwidth costs increase (by increasing batch size), the speedups decrease. This suggests that CA-SGD might perform even better on cloud/commodity resources. Implementing and benchmarking CA-SGD on cloud resources and programming models would be very interesting. %Additional performance tuning of the Gram matrix kernel and exploring reduced-precision storage formats (where bandwidth/flops costs decrease but latency cost is fixed) would also make CA-SGD more impactful.
\subsubsection{Implications for neural networks}
Backpropagation in neural networks introduces a set of nested recurrence relations with nonlinear activation functions at each layer and hidden unit. Since logistic regression can be interpreted as a single-layer neural network, we can likely apply our technique to feedforward neural networks (FNN) with nonlinear activation functions. The extension to convolution layers should also be straighforward given that convolutions are linear operations. However, for practical applications to CNNs we need to assess whether the $s$-step derivation can be applied to batch normalization and pooling layers. While we believe that the $s$-step technique can be extended to FNNs and CNNs, hand deriving CA-variants for each individual FNN/CNN model is unscalable. Therefore, it is critical to develop tools and techniques that can help automate the CA-derivation process. Finally, as models get wider and deeper, they become more compute and bandwidth bound. This suggests that our approach is most impactful when large models are scaled out to a latency-bound setting.
\ifCLASSOPTIONcompsoc
  \section*{Acknowledgments}
\else
  \section*{Acknowledgment}
\fi
We would like to thank the anonymous reviewers for their helpful feedback. Computational resources were provided by the Maryland Advanced Research Computing Center. AD is supported by the Gordon and Betty Moore Foundation.% under JHU grant no. 130764.
%\textcolor{red}{\bf To be completed.}

\ifCLASSOPTIONcaptionsoff
  \newpage
\fi

%\IEEEtriggeratref{9}


% BibTeX documentation can be easily obtained at:%
\bibliographystyle{IEEEtran}
\bibliography{refs}
%\nocite{*}
%\begin{thebibliography}{1}
%\end{thebibliography}


\begin{IEEEbiography}{Aditya Devarakonda}received a B.S. in Computer Engineering from Rutgers University, New Brunswick in 2012, an M.S. and Ph.D. in Computer Science from the University of California at Berkeley in 2016 and 2018, respectively. He is currently an assistant research scientist in the Department of Physics and Astronomy at the Johns Hopkins University. His research interests include the theory and practice of parallel machine learning and applications of machine learning to other scientific domains. He is a member of the ACM, the IEEE, and the SIAM.
\end{IEEEbiography}

\begin{IEEEbiography}{James Demmel}
\textcolor{red}{\textbf{Add bio.}}
\end{IEEEbiography}


%\newpage


%\vfill


\end{document}